\documentclass[12pt]{article}

\usepackage{scicite}
\usepackage{times}

\usepackage{amsmath}%
\usepackage{amsfonts}%
\usepackage{amssymb}%
\usepackage{mathrsfs}
\usepackage[capitalize]{cleveref}
\usepackage{overpic}
\usepackage{url}
\newcommand{\R}{\mathbb{R}}
\renewcommand{\d}{\,\textup{d}}
\renewcommand{\L}{\mathcal{L}}
\newcommand{\D}{\mathcal{D}}
\newcommand{\N}{\mathcal{N}}

\usepackage{xcolor}

\newenvironment{sciabstract}{%
\begin{quote} \bf}
{\end{quote}}

\title{Data-driven discovery of Green's functions with human-understandable deep learning}

\author
{Nicolas Boull\'e,$^{1}$ Christopher J. Earls,$^{2,3}$ Alex Townsend$^{2,4}$\\
\\
\normalsize{$^{1}$Mathematical Institute, University of Oxford, Oxford OX2 6GG, UK}\\
\normalsize{$^{2}$Center for Applied Mathematics, Cornell University, Ithaca, NY 14853, USA}\\
\normalsize{$^{3}$School of Civil and Environmental Engineering, Cornell University, Ithaca, NY 14853, USA}\\
\normalsize{$^{4}$Department of Mathematics, Cornell University, Ithaca, NY 14853, USA}\\
}

\date{}

\begin{document}

\baselineskip24pt

\maketitle 

\begin{sciabstract}
There is an opportunity for deep learning to revolutionize science and technology by revealing its findings in a human interpretable manner. To do this, we develop a novel data-driven approach for creating a human-machine partnership to accelerate scientific discovery. By collecting physical system responses under excitations drawn from a Gaussian process, we train rational neural networks to learn Green's functions of hidden linear partial differential equations. These functions reveal human-understandable properties and features, such as linear conservation laws and symmetries, along with shock and singularity locations, boundary effects, and dominant modes. We illustrate the technique on several examples and capture a range of physics, including advection-diffusion, viscous shocks, and Stokes flow in a lid-driven cavity.
\end{sciabstract}

Deep learning (DL) holds promise as a scientific tool for discovering elusive patterns within the natural and technological world~\cite{lecun2015deep,goodfellow2016deep}. These patterns hint at undiscovered partial differential equations (PDEs) that describe governing phenomena within biology, fluid dynamics, and physics. From sparse and noisy laboratory observations, we aim to learn mechanistic laws of nature~\cite{brunton2020machine,karniadakis2021physics}. Recently, scientific computing and machine learning have successfully converged on PDE discovery~\cite{Brunton,schaeffer2017learning,Rudy,zhang2020data}, PDE learning~\cite{raissi2018deep,lu2021learning,gin2020deepgreen,li2020neural,feliu2020meta,raissi2020hidden}, and symbolic regression~\cite{schmidt2009distilling,Udrescu2020} as promising means for applying machine learning to scientific investigations. These methods attempt to discover the coefficients of a PDE model or learn the operator that maps excitations to system responses. The recent DL techniques addressing the latter problem are based on approximating the solution operator associated with a PDE by a neural network~\cite{raissi2018deep,lu2021learning,gin2020deepgreen,li2020neural,feliu2020meta}. While excellent for solving PDEs, we consider them as ``black box'' and focus here on a data-driven strategy that improves human understanding of the governing PDE model.

In contrast, we offer a radically different, alternative approach that is backed by theory~\cite{boulle2021learning} and infuse an interpretation in the model by learning well-understood mathematical objects that imply underlying physical laws. We devise a DL method, employed for learning the Green's functions~\cite{stakgold2011green} associated with unknown governing linear PDEs, and train the neural networks (NNs) by collecting physical system responses from random excitation functions drawn from a Gaussian process. The empirically derived Green's functions relate the system's response (or PDE solution) to a forcing term, and can then be used as a fast reduced-order PDE solver. The existing Graph Kernel Network~\cite{li2020neural} and DeepGreen~\cite{gin2020deepgreen} techniques also aim to learn solution operators of PDEs based on Green's functions. While they show competitive performance in predicting the solution of the PDE for new forcing functions, they fail to capture Green's functions accurately, which makes the extraction of qualitative and quantitative features of the physical system challenging.

Our primary objective is to study the discovered Green's functions for clues regarding the physical properties of the observed systems. Our approach relies on a novel and adaptive neural network architecture called a rational neural network~\cite{boulle2020rational}, which has higher approximation power than standard networks and carries human-understandable features of the PDE, such as shock and singularity locations.

In this paper, we use techniques from deep learning to discover the Green's function of linear differential equations $\L u = f$ from input-output pairs $(f,u)$, as opposed to directly learning $\L$, or model parameters.
In this sense, our approach is agnostic to the forward PDE model, but nonetheless offers insights into its physical properties. There are several advantages to learning the Green's function. First, once the Green's function is learned by a neural network (NN), it is possible to compute the solution, $u$, for a new forcing term, $f$, by evaluating an integral (see~\cref{eq_Green_integral}); which is more efficient than training a new NN. Second, the Green's function associated with $\mathcal{L}$ contains information about the operator, $\mathcal{L}$, and the type of boundary constraints that are imposed; which helps uncover mechanistic understanding from experimental data. Finally, it is easier to train NNs to approximate Green's functions, which are square-integrable functions under sufficient regularity conditions~\cite{stakgold2011green,gruter1982green,dong2009green}, than trying to approximate the action of the linear differential operator, $\L$, which is not bounded~\cite{Kreyszig}. Also, any prior mathematical and physical knowledge of the operator, $\L$, can be exploited in the design of the NN architecture, which could enforce a particular structure such as symmetry of the Green's function.

\section*{Results}
\subsection*{Deep learning Green's functions.}

\begin{figure*}[ht!]
\centering
\vspace{0.5cm}
\begin{overpic}[width=\textwidth, trim=0 0 0 0,clip]{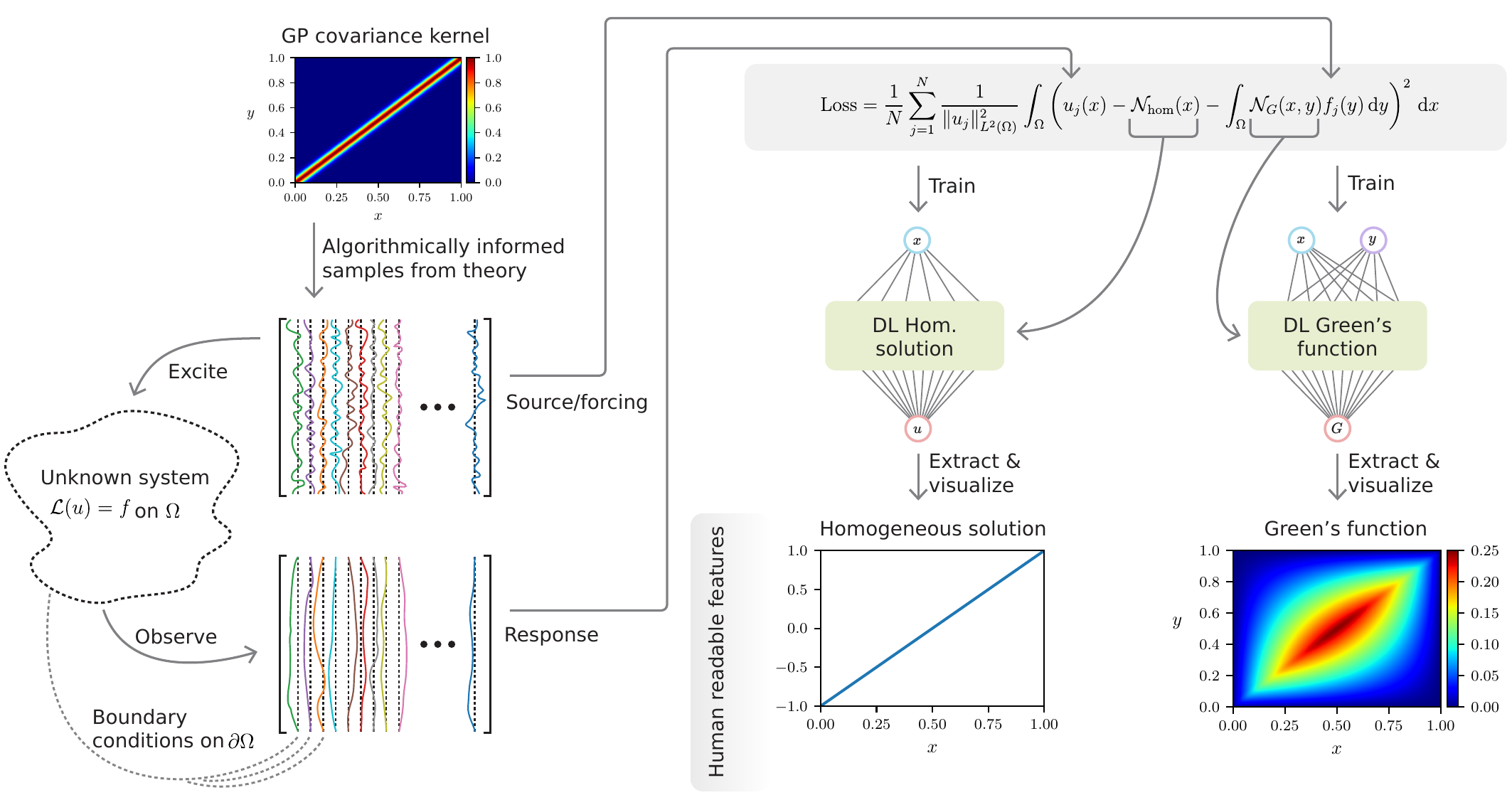}
\put(15,50){\textbf{A}}
\put(16,33){\textbf{B}}
\put(16,17){\textbf{C}}
\put(46,47){\textbf{D}}
\put(51,38){\textbf{E}}
\put(51,20){\textbf{F}}
\end{overpic}
\caption{Schematic of our DL method for learning Green's functions from input-output pairs.  (A) The covariance kernel of the Gaussian process (GP), which is used to generate excitations. (B) The random excitations and the system's response are recorded (C). (D) A loss function is minimized to train rational NNs (E). (F) The learned Green's function and homogeneous solution are visualized by sampling the NNs.}
\label{fig_idea}
\end{figure*}
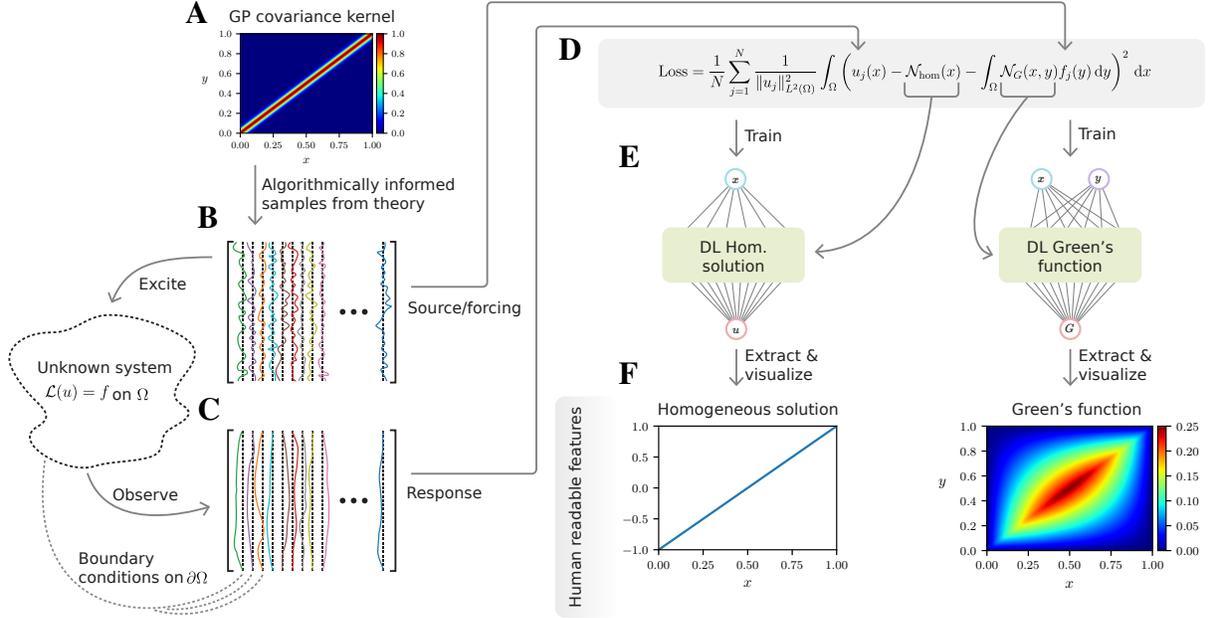

Our DL approach (see~\cref{fig_idea}) begins with excitations (or forcing terms), $\{f_j\}_{j=1}^N$, sampled from a Gaussian process (GP) having a carefully designed covariance kernel~\cite{boulle2021learning}, and corresponding system responses, $\{u_j\}_{j=1}^N$. It is postulated that there is an unknown linearized governing PDE so that $\L u_j = f_j$. The selection of random forcing terms is theoretically justified~\cite{boulle2021learning} and enables us to learn the dominant eigenmodes of the solution operator, using only a small number, $N$, of training pairs. The Green's function, $G$, and homogeneous solution, $u_{\text{hom}}$, which encodes the boundary conditions associated with the PDE, satisfy
\begin{equation} \label{eq_Green_integral}
u_j(x) = \int_{\Omega}G(x,y)f_j(y)\d y + u_{\text{hom}}(x),\qquad x\in\Omega,
\end{equation}
and are approximated by two rational neural networks: $\N_G$ and $\N_{\text{hom}}$.

A rational NN consists of a NN with trainable rational activation functions whose coefficients are learned simultaneously with the weights and biases of the network. Rational NNs have better approximation properties than standard NNs~\cite{boulle2020rational}, both in theory and in practice, which makes them ideal for the present application. The parameters of the NNs representing the Green's function and homogeneous solution are simultaneously learned through minimization of the loss function displayed in \cref{fig_idea}D (Supplementary Material, \cref{sec_loss}). We discretize the integrals in the loss function at the specified measurement locations $\{x_i\}_{i=1}^{N_u}$, within the domain, $\Omega$, and forcing term sample points, $\{y_i\}_{i=1}^{N_f}$, respectively, using a quadrature rule. 

In the Supplementary Material, \cref{fig_robustness}, we also present numerical results obtained from sparse training data, or noisy spatial measurements, which demonstrate the robustness of our method (Supplementary Material, \cref{sec_robustness}). Additionally, our DL technique is data-driven and requires minimal by-hand parameter tuning. In fact, all the numerical examples described here and in the Supplementary Material are performed using a unique rational NN architecture, initialization procedure, and optimization algorithm.

\subsection*{Human-understandable features.}

\begin{figure*}[t]
\centering
\vspace{0.5cm}
\begin{overpic}[width=\textwidth, trim=0 0 0 0,clip]{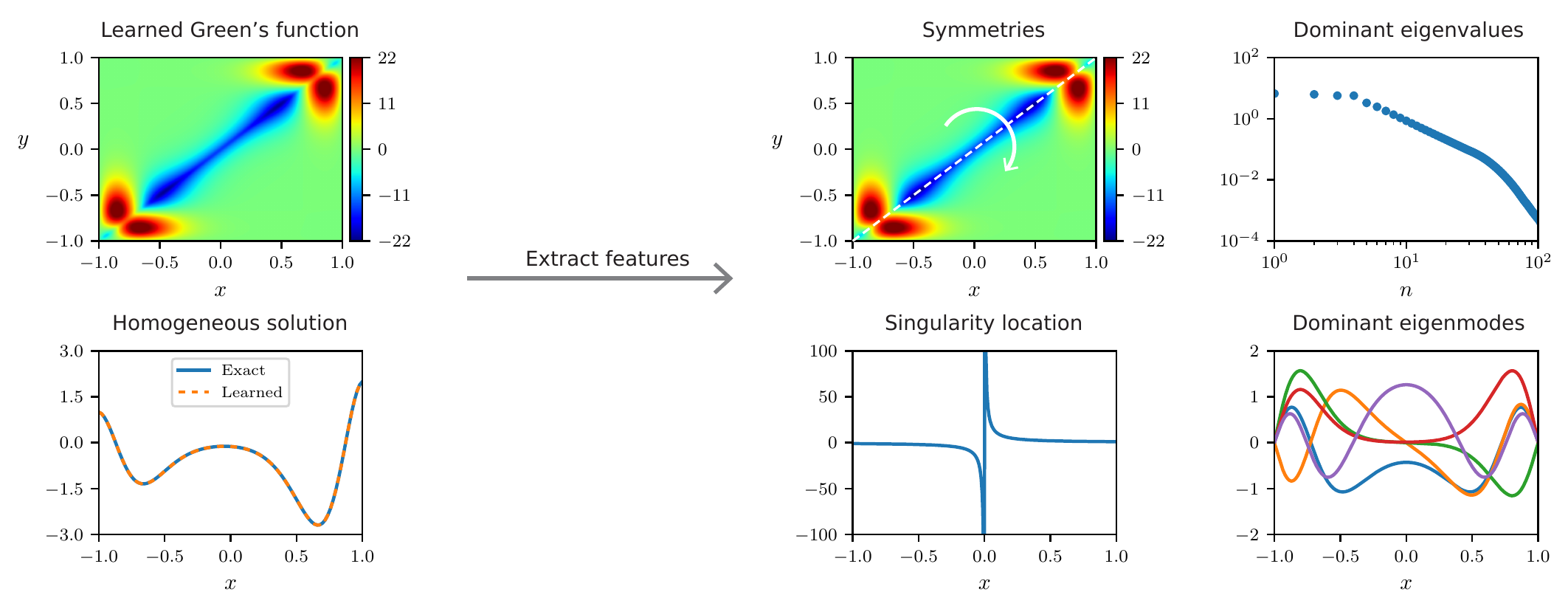}
\put(0.5,37){\textbf{A}}
\put(0.5,18){\textbf{B}}
\put(48,37){\textbf{C}}
\put(48,18){\textbf{D}}
\put(75,37){\textbf{E}}
\put(75,18){\textbf{F}}
\end{overpic}
\caption{Feature extraction from learned Green's functions. The NNs for the learned Green's function (A) and homogeneous solution (B) enable the extraction of qualitative and quantitative features associated with the differential operator. For example, the  symmetries in the Green's function reveal PDE invariances (C), poles of rational NNs identify singularity type and location (D), the dominant eigenvalues (E) and eigenmodes (F) of the learned Green's function are related to the eigenvalues and eigenmodes of the differential operator.}
\label{fig_schema}
\end{figure*}
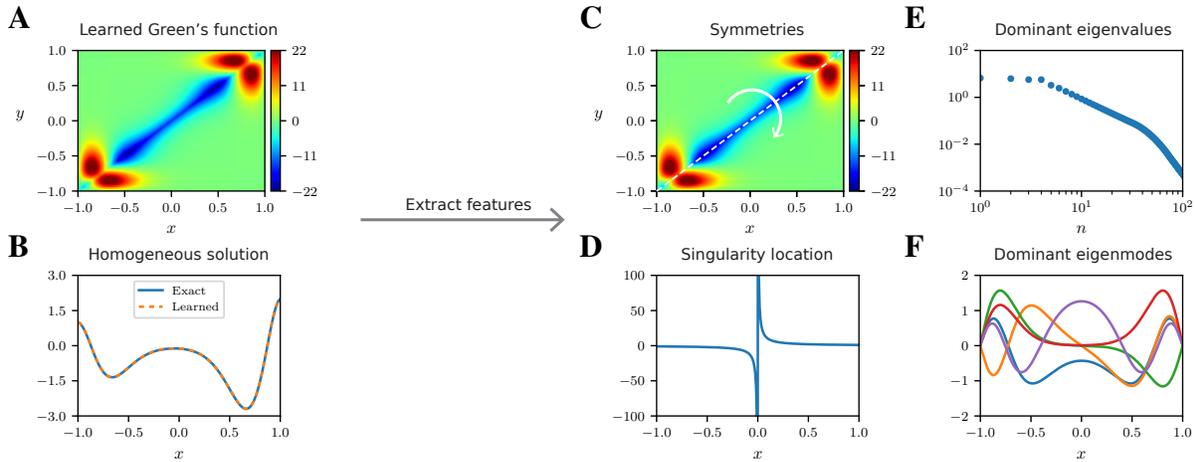

The trained NNs contain both the desired Green's function and homogeneous solution, which we evaluate and visualize to glean novel insights concerning the underlying, governing PDE (\cref{fig_schema}). In this way, we achieve one part of our human interpretation goal: finding a link between the properties of the Green's function and that of the underlying differential operator and solution constraints. 

As an example, if the Green's function is symmetric, i.e., $G(x,y)=G(y,x)$ for all $x,y\in\Omega$, then the operator $\L$ is self-adjoint. Another aspect of human interpretability is that the poles of the trained rational NN tend to cluster in a way that reveal the location and type of singularities in the homogeneous solution. Finally, there is a direct correspondence between the dominant eigenmodes and eigenvalues (as well as the singular vectors and singular values) of the learned Green's function and those of the differential operator. The correspondence gives insight into the important eigenmodes that govern the system's behavior (Supplementary Material, \cref{sec_features}).

\begin{figure*}[ht!]
\centering
\vspace{0.5cm}
\begin{overpic}[width=\textwidth]{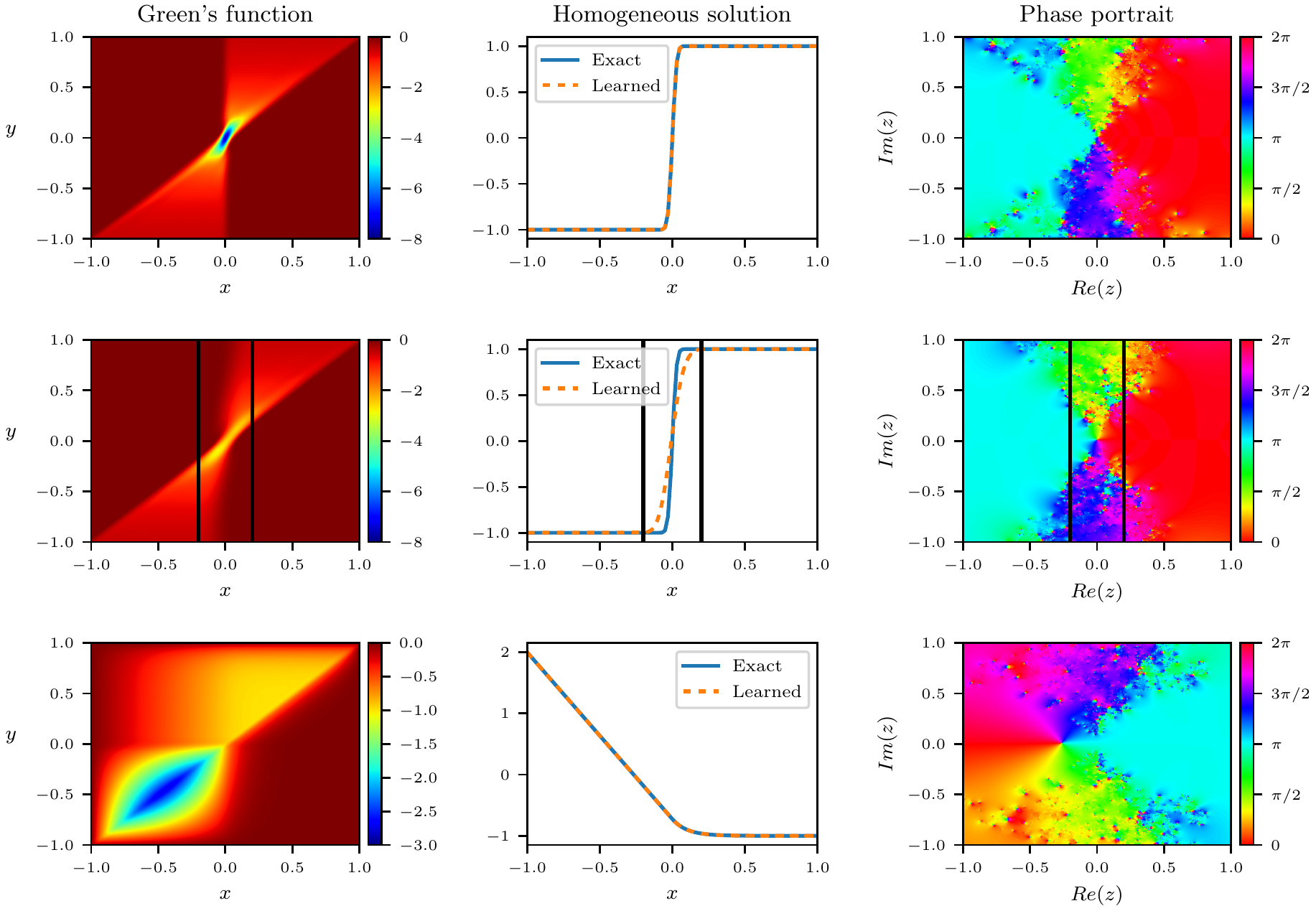}
\put(0,67){\textbf{A}}
\put(34,67){\textbf{B}}
\put(65,67){\textbf{C}}
\put(0,44){\textbf{D}}
\put(34,44){\textbf{E}}
\put(65,44){\textbf{F}}
\put(0,21){\textbf{G}}
\put(34,21){\textbf{H}}
\put(65,21){\textbf{I}}
\end{overpic}
\caption{Green's functions learned by rational neural networks. (A) Green's function of a differential operator with a viscous shock at $x=0$, learned by a rational NN. (B) Learned and exact (computed by a classical spectral method) homogeneous solution to the differential equation with zero forcing term. (C) Phase portrait of the homogeneous rational NN evaluated on the complex plane. (D to F) Similar to (A to C), but without any system's response measurements in $x\in[-0.2,0.2]$ (see vertical black lines) near the shock. (G) Learned Green's function and homogeneous solution (H) of an advection-diffusion operator with advection occurring for $x\geq 0$. (I) Phase portrait of the homogeneous NN on the complex plane.}
\label{fig_experiments}
\end{figure*}
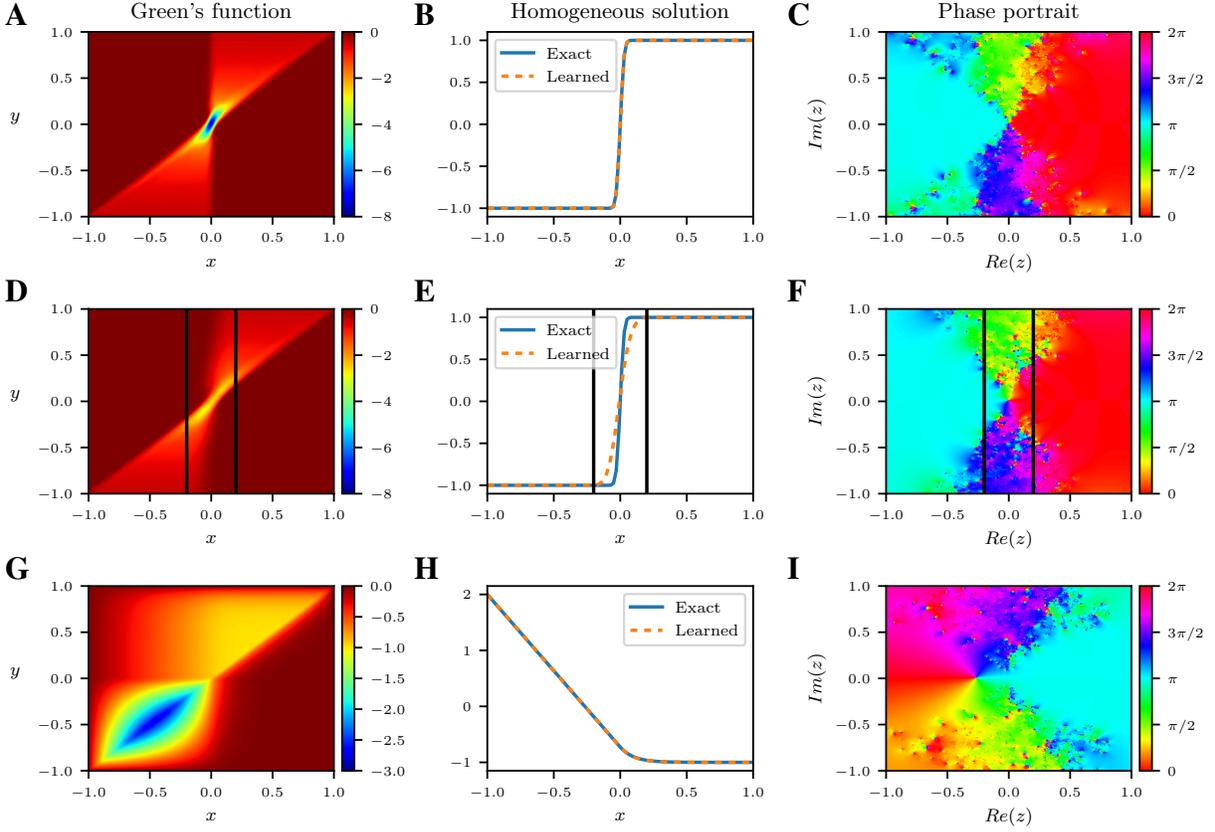

\subsection*{Numerical examples.}

As a first example, we consider a second-order differential operator having suitable variable coefficients to model a viscous shock at $x=0$~\cite{lee1997fast}. The system's responses are obtained by solving the PDE, with Dirichlet boundary conditions, using a spectral numerical solver for each of the $N=100$ random forcing terms, sampled from a GP having a squared-exponential covariance kernel~\cite{boulle2021learning}. The learned Green's function is displayed in \cref{fig_experiments}A and satisfies the following symmetry relation: $G(x,y) = G(-x,-y)$, indicating the presence of a reflective symmetry group within the underlying PDE. Indeed, if $u$ is a solution to $\L u=f$ with homogeneous boundary conditions, then $u(-x)$ is a solution to $\L v=f(-x)$. We also observe in \cref{fig_experiments}B and C that the homogeneous solution is accurately captured and that the poles of the homogeneous rational NN cluster near the real axis around $x=0$: the location of the singularity induced by the shock (Supplementary Material, \cref{sec_singularity}). 

Next, we reproduce the same viscous shock numerical experiment, except that this time we remove measurements of the system's response from the training dataset in the interval $[-0.2,0.2]$: adjacent to the shock front. By comparing \cref{fig_experiments}D to F and \cref{fig_experiments}A to C, we find that the Green's function and homogeneous solution, learned by the rational NNs, may not be affected in the region outside of the interval with missing data. In some cases, the NNs can still accurately capture the main features of the Green's function and homogeneous solution in the region lacking measurements. The robustness of our method to noise perturbation and corrupted or missing data is of significant interest and promising for real applications with experimental data.

We next apply our DL method to discover the Green's function and homogeneous solution of an advection-diffusion operator, where the advection is dominant only within the right half of the domain. The output of the Green's function NN is plotted in \cref{fig_experiments}G, where we observe the disparate spatial behaviors of the dominant physical mechanisms. This can be recognized when observing the restriction of the Green's function to the subdomain $[-1,0]\times[-1,0]$, where the observed solution is reminiscent of the Green's function for the Laplacian; thus indicating that the PDE is diffusive on the left half of the domain. Similarly, the restriction of the learned Green's function to $[0,1]\times[0,1]$ is characteristic of advection. 

In \cref{fig_experiments}H and I, we display the homogeneous solution NN, along with the phase of the rational NN, evaluated on the complex plane. The agreement between the exact and learned homogeneous solution illustrates the ability of the DL method to accurately capture the behavior of a system within ``multiphysics'' contexts. The choice of rational NNs is crucial here: to deepen our understanding of the system, as the poles of the homogeneous rational NN characterize the location and type of singularities in the homogeneous solution. Here the change in behavior of the differential operator from diffusion to advection is delineated by the location of the poles of the rational NN. 

\subsection*{Nonlinear and vector-valued equations.}

\begin{figure*}[ht]
\centering
\vspace{0.5cm}
\begin{overpic}[width=\textwidth]{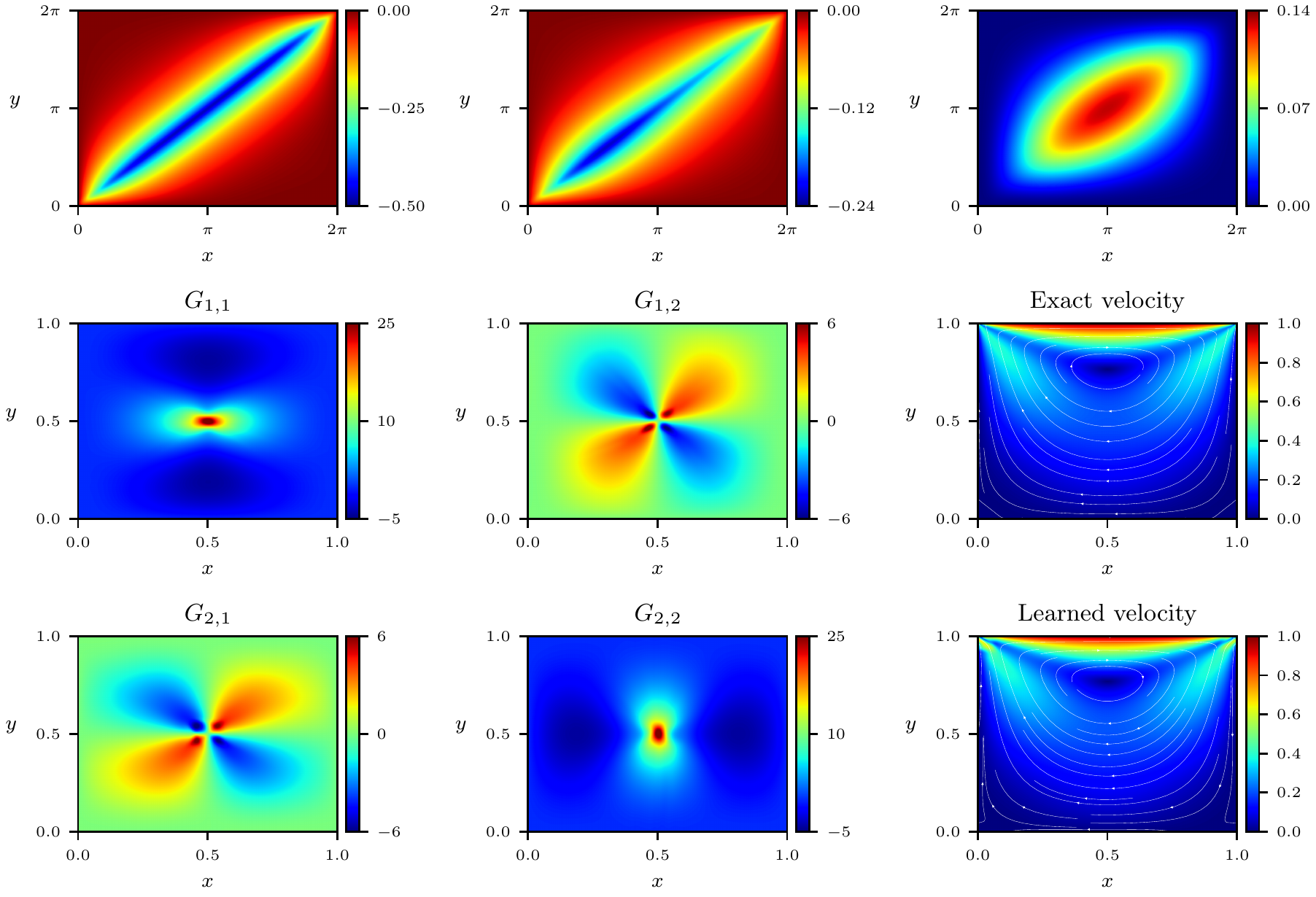}
\put(0,67.5){\textbf{A}}
\put(34,67.5){\textbf{B}}
\put(68,67.5){\textbf{C}}
\put(0,44){\textbf{D}}
\put(68,44){\textbf{E}}
\put(68,21){\textbf{F}}
\end{overpic}
\caption{Linearized models and Stokes flow. (A to C) Green's functions of three differential operators: Helmholtz, Sturm--Liouville, and biharmonic, with cubic nonlinearity. (D) Matrix of Green's functions of a two-dimensional Stokes flow in a lid-driven cavity, evaluated at a two-dimensional slice. Velocity magnitude and streamlines of the exact (E) and learned (F) homogeneous solution to the Stokes equations with zero applied body force.}
\label{fig_stokes}
\end{figure*}
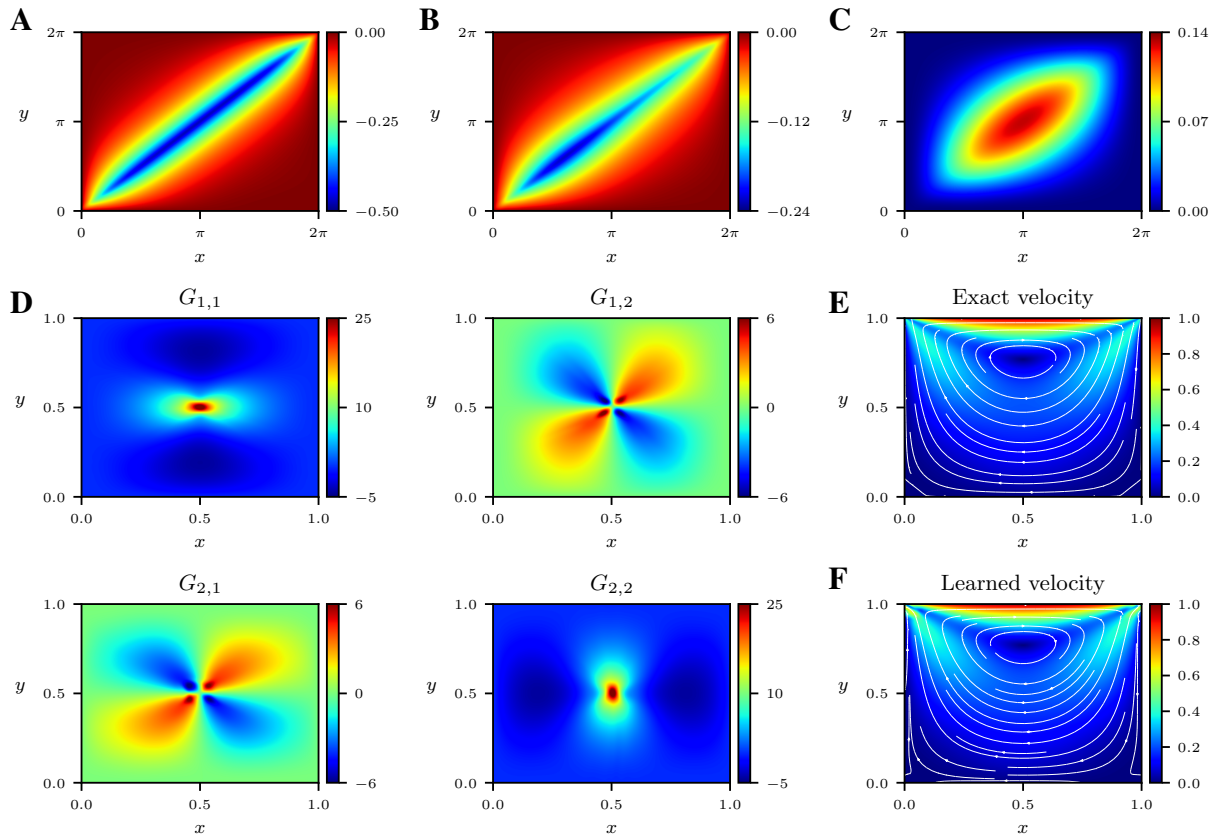

We can also discover Green's functions from forcing terms and concomitant solutions to nonlinear differential equations possessing semi-dominant linearity. In \cref{fig_stokes}A to C, we visualize the Green's function NNs of three operators with cubic nonlinearity considered in~\cite{gin2020deepgreen}. The nonlinearity does not prevent our method from discovering a Green's function of an approximate linear model, from which one can understand features such as symmetry and boundary conditions. This property is crucial for tackling time-dependent problems, where the present technique may be extended and applied to uncover linear propagators.

Finally, we consider a Stokes flow in a two-dimensional lid-driven cavity to emphasize the ability of our method to handle systems of differential equations in two dimensions. In this context, the relation between the system's responses and the forcing terms can be expressed using a Green's matrix, which consists of a two-by-two matrix of Green's functions and whose components reveal features of the underlying system such as symmetry and coupling (\cref{fig_stokes}D and the Supplementary Material, \cref{sec_system,sec_lid_driven}). \cref{fig_stokes}E and F illustrate that the homogeneous solution to the Stokes equation is accurately captured by the homogeneous rational NN, despite the corner singularities and coarse measurement grid (Supplementary Material, \cref{sec_lid_driven}).

\section*{Discussion}

Contrary to existing works in the literature~\cite{gin2020deepgreen,lu2021learning,li2020neural,li2020fourier}, our primary aim is to uncover mechanistic understanding from input-output data using a human-understandable representation of the underlying, but hidden, differential operator. This representation takes the form of a rational NN~\cite{boulle2020rational} for the Green's function. We extensively described all the physical features of the operator that can be extracted and discovered from the learned Green's function and homogeneous solutions, such as linear conservation laws, symmetries, shock front and singularity locations, boundary conditions, and dominant modes.

The DL method for learning Green's functions of linear differential operators naturally extends to the case of three spatial dimensions but these systems are more challenging due to the GPU memory demands required to represent the six-dimensional inputs used to train the NN representing the Green's function. However, alternative optimization algorithms than the one used in this paper and described in \emph{Methods}, such as mini-batch optimization~\cite{kingma2014adam,li2014efficient}, may be employed to alleviate the computational expense of the training procedure.

While our method is demonstrated on linear differential operators, it can be extended to nonlinear, time-dependent problems that can be linearized using an implicit-explicit time-stepping scheme~\cite{ascher1997implicit,pareschi2005implicit} or an iterative method~\cite{kelley1995iterative}. This process allows us to learn the Green's functions of linear time propagators and understand physical behavior in time-dependent problems from input-output data such as the time-dependent Schr\"odinger equation (Supplementary Material, \cref{sec_time_dep}). The numerical experiments conducted in \cref{fig_stokes}A to C highlight that our approach can discover Green's functions of linearizations of nonlinear differential operators.

Our deep learning method for learning Green's functions and extracting human-understandable properties of partial differential equations benefits from the adaptivity of rational neural networks and its support for qualitative feature detection and interpretation. We successfully tested our approach with noisy and sparse measurements as training data (Supplementary Material, \cref{sec_robustness}). The design of our applied network architectures, and covariance kernel used to generate the system forcing is guided by rigorous theoretical statements~\cite{boulle2021learning,boulle2020rational} that offer performance guarantees. This shows that our proposed deep learning method may be used to discover new mechanistic understanding with machine learning.

\section*{Methods}

\subsection*{Green's functions.} 
We consider linear differential operators, $\L$, defined on a bounded domain $\Omega\subset\R^d$, where $d\in\{1,2,3\}$ denotes the spatial dimension. The aim of our method is to discover properties of the operator, $\L$, using $N$ input-output pairs $\{(f_j,u_j)\}_{j=1}^N$, consisting of forcing functions, $f_j:\Omega\to\R$, and system responses, $u_j:\Omega\to\R$, which are solutions to the following equation:
\begin{equation} \label{eq_problem}
\L u_j = f_j, \qquad \mathcal{D}(u_j,\Omega) = g,
\end{equation}
where $\mathcal{D}$ is a linear operator acting on the solutions, $u$, and the domain, $\Omega$; with $g$ being the constraint. We assume that the forcing terms have sufficient regularity, and that the operator, $\D$, is a constraint so that \cref{eq_problem} has a unique solution~\cite{stakgold2011green}. An example of constraint is the imposition of homogeneous Dirichlet boundary conditions on the solutions: $\mathcal{D}(u_j,\Omega) := u_{j}|_{\partial\Omega}=0$. Note that boundary conditions, integral conditions, jump conditions, or non-standard constraints, are all possible (Supplementary Material, \cref{sec_linear_const}). 

A Green's function~\cite{evans10,arfken2011mathematical,myint2007linear,stakgold2011green} of the operator, $\L$, is defined as the solution to the following equation:
\[\L G(x,y) = \delta(y-x),\qquad x,y\in\Omega,\] 
where $\L$ is acting on the function $x\mapsto G(x,y)$ for fixed $y\in\Omega$, and $\delta(\cdot)$ denotes the Dirac delta function. The Green's function is well-defined and unique under mild conditions on $\L$, and suitable solution constraints imposed via an operator, $\mathcal{D}$ (see~\cref{eq_problem})~\cite{stakgold2011green}. Moreover, if $(f,u)$ is an input-output pair, satisfying \cref{eq_problem} with $g=0$, then
\[u(x) = \int_{\Omega}G(x,y)f(y)\d y, \qquad x\in\Omega.\] 
Therefore, the Green's function associated with $\mathcal{L}$ can be thought of as the right inverse of $\mathcal{L}$. 

Let $u_{\text{hom}}$ be the homogeneous solution to~\eqref{eq_problem}, so that
\[\L u_{\text{hom}} = 0, \qquad \mathcal{D}(u_{\text{hom}},\Omega) = g.\]  
Using superposition, we can construct solutions, $u_j$, to \cref{eq_problem} as $u_j = \tilde{u}_j+u_{\text{hom}}$, where $\tilde{u}_j$ satisfies
\[\L \tilde{u}_j = f_j, \qquad \mathcal{D}(\tilde{u}_j,\Omega) = 0.\] 
Then, the relation between the system's response, $u_j$, and the forcing term, $f_j$, can be expressed via the Green's function as
\[u_j(x) = \int_{\Omega}G(x,y)f_j(y)\d y + u_{\text{hom}}(x),\qquad x\in\Omega.\] 
Therefore, we train two NNs: $\N_G:\Omega\times\Omega\to\R\cup\{\pm\infty\}$ and $\N_{\text{hom}}:\Omega\to\R$, to learn the Green's function, and also the homogeneous solution associated with $\L$ and the constraint operator $\mathcal{D}$. Note that this procedure allows us to discover boundary conditions, or constraints, directly from the input-output data without imposing it in the loss function (which often results in training instabilities~\cite{wight2020solving}).

\subsection*{Rational neural networks.} \label{sec_rational_net}

Rational NNs~\cite{boulle2020rational} consist of NNs with adaptive rational activation functions $x\mapsto\sigma(x) = p(x)/q(x)$, where $p$ and $q$ are two polynomials, whose coefficients are trained at the same time as the other parameters of the networks, such as the weights and biases. These coefficients are shared between all the neurons in a given layer but generally differ between the network's layers. This type of network was proven to have better approximation power than standard Rectified Linear Unit (ReLU) networks~\cite{glorot2011deep,yarotsky2017error}, which means that they can approximate smooth functions more accurately with fewer layers and network parameters~\cite{boulle2020rational}. It is also observed in practice that rational NNs require fewer optimization steps and therefore can be more efficient to train than other activation functions~\cite{boulle2020rational}.

The NNs, $\N_G$ and $\N_{\text{hom}}$, which approximate the Green's function and homogeneous solution associated with \cref{eq_problem}, respectively, are chosen to be rational NNs~\cite{boulle2020rational} with 4 hidden layers and 50 neurons in each layer. We choose the polynomials, $p$ and $q$, within the activation functions to be of degree 3 and 2, respectively, and initialize the coefficients of all the rational activation functions so that they are the best $(3,2)$ rational approximant to a ReLU (see the supplementary material of~\cite{boulle2020rational} for details). The motivation is that the flexibility of the rational functions brings extra benefit in the training and accuracy over the ReLU activation function. We highlight that the increase in the number of trainable parameters, due to the adaptive rational activation functions, is only linear with respect to the number of layers and negligible compared to the total number of parameters in the network as:
\[\text{number of rational coefficients} = 7\times \text{number of hidden layers} = 28.\]
The weight matrices of the NNs are initialized using Glorot normal initializer~\cite{glorot2010understanding}, while the biases are initialized to zero.

Another advantage of rational NNs is the potential presence of poles, \emph{i.e.}, zeros of the polynomial $q$. While the initialization of the activation functions avoids training issues due to potential spurious poles, the poles can be exploited to learn physical features of the differential operator (Supplementary Material, \cref{sec_singularity}). Therefore, the architecture of the NNs also supports the aim of a human-understandable approach for learning PDEs. In higher dimensions, such as $d = 2$ or $d =3$, the Green's function is not necessarily bounded along the diagonal, i.e., $\{(x,x),\, x\in\Omega\}$; thus making the poles of the rational NNs crucial.

Finally, we emphasize that the enhanced approximation properties of rational NNs~\cite{boulle2020rational} make them ideal for learning Green's functions and, more generally, approximating functions within regression problems. These networks may also be of benefit to other approaches for solving and learning PDEs with DL techniques, such as PINNs~\cite{raissi2019physics}, DeepGreen~\cite{gin2020deepgreen}, DeepONet~\cite{lu2021learning}, Neural operator~\cite{li2020neural}, and Fourier neural operator~\cite{li2020fourier}. 

\subsection*{Data generation.}

We create a training dataset, consisting of input-output functions, $\{(f_j\,u_j)\}$ for $1\leq j \leq N$, in three steps: (1) Generating the forcing terms by sampling random functions from a Gaussian process (GP), (2) Solving \cref{eq_problem} for the generated forcing terms, and (3) Sampling the forcing terms, $f_j$, at the points $\{y_1,\ldots,y_{N_f}\}\subset\Omega$ and the system's responses, $u_j$, at $\{x_1,\ldots,x_{N_u}\}\subset\Omega$. Here, $N_f$ and $N_u$ are the forcing and solution discretization sizes, respectively. We recommend that all the forcing terms are sampled on the same grid and similarly for the system's responses. This minimizes the number of evaluations of $\N_G$ during the training phase and reduces the computational and memory costs of training. 

The spatial locations of points $\{y_i\}$ and the forcing discretization size, $N_f$, are chosen arbitrarily to train the NNs as the forcing terms are assumed to be known over $\Omega$. In practice, the number, $N_u$, and location of the measurement points, $\{x_i\}$, are imposed by the nature of the experiment, or simulation, performed to measure the system's response. When $\Omega$ is an interval, we always select $N_f=200$, $N_u=100$, and equally-spaced sampled points for the forcing and response functions. Further details on the training data generation are available in the Supplementary Material, \cref{sec_generation_data}. We then analyze the robustness of our method for learning Green's functions with respect to the number and location of the measurement points in the Supplementary Material, \cref{sec_robustness}.

\subsection*{Neural network training.}

The NNs are implemented with single-precision floating-point format within the TensorFlow DL library~\cite{abadi2016tensorflow}, and are trained (the numerical experiments are performed on a desktop computer with a Intel\textsuperscript{\tiny\textregistered} Xeon\textsuperscript{\tiny\textregistered} CPU E5-2667 v2 @ 3.30GHz and a NVIDIA\textsuperscript{\tiny\textregistered} Tesla\textsuperscript{\tiny\textregistered} K40m GPU) using a two-step optimization procedure to minimize the loss function (Supplementary Material, \cref{sec_loss}). First, we use Adam's algorithm~\cite{kingma2014adam} for the first 1000 optimization steps (or epochs), with default learning rate $0.001$ and parameters $\beta_1 = 0.9$, $\beta_2 = 0.999$. Then, we employ the limited memory BFGS, with bound constraints (L-BFGS-B) optimization algorithm~\cite{liu1989limited,byrd1995limited}, implemented in the SciPy library~\cite{virtanen2020scipy}, with a maximum of $5\times 10^4$ iterations. This training procedure is used by Lu \emph{et al.} to train physics-informed NNs (PINNs) and mitigate the risk of the optimizer getting stuck at a poor local minima~\cite{lu2021deepxde}. The L-BFGS-B algorithm is also successful for PDE learning~\cite{raissi2018deep} and PDE solvers using DL techniques~\cite{raissi2019physics,lu2021deepxde}. Moreover, this optimization algorithm takes advantage of the smoothness of the loss function by using second-order derivatives and often converges in fewer iterations than Adam's algorithm and other methods based on stochastic gradient descent~\cite{lu2021deepxde}. Within this setting, rational NNs are beneficial because the activation functions are smooth while maintaining an initialization close to ReLU (Supplementary Material, \cref{fig_loss}).

\subsection*{Theoretical justification.}
Our approach for learning Green's functions associated with linear differential operators has a theoretically rigorous underpinning. Indeed, it was shown in~\cite{boulle2021learning} that uniformly elliptic operators in three dimensions have an intrinsic \emph{learning rate}, which characterizes the number of training pairs needed to construct an $\epsilon$-approximation in the $L^2$-norm of the Green's function, $G$, with high probability, for $0<\epsilon<1$. The number of training pairs depends on the quality of the covariance kernel used to generate the random forcing terms, $\{f_j\}_{j=1}^N$. Our choice of covariance kernel (Supplementary Material, section~1) is motivated by the GP quality measure~\cite{boulle2021learning,boulle2021generalization}, to ensure that our set of training forcing terms is sufficiently diverse to capture the action of the solution operator, $f\mapsto u(x) = \int_{\Omega}G(x,y)f(y)\d y$, on a diverse set of functions.

Similarly, the choice of rational NNs to approximate the Green's function, and the homogeneous solution, is justified by the higher approximation power of these networks over ReLU~\cite{boulle2020rational}. Other adaptive activation functions have been proposed for learning or solving PDEs with NNs~\cite{jagtap2020adaptive}, but they are only motivated by empirical observations. Both theory and experiments support rational NNs for regression problems. The number of trainable parameters, consisting of weight matrices, bias vectors, and rational coefficients, needed by a rational NN to approximate smooth functions within $0<\epsilon<1$, can be completely characterized~\cite{boulle2020rational}. This motivates our choice of NN architecture for learning the Green's functions.

\section*{Data Availability}
All data and codes used in this article and the Supplementary Material are publicly available on the GitHub and Zenodo repositories at \url{https://github.com/NBoulle/greenlearning/}~\cite{boulleZenodo} to reproduce the numerical experiments and figures. A software package, including additional examples and documentation, is also available at  \url{https://greenlearning.readthedocs.io/}.

\section*{Acknowledgments}
We thank Gregory Bewley for suggestions on the manuscript. This work was supported by the EPSRC Centre for Doctoral Training in Industrially Focused Mathematical Modelling through grant EP/L015803/1 in collaboration with Simula Research Laboratory. C.J.E. was supported by the Army Research Office (ARO) Biomathematics Program grant W911NF-18-1-0351. A.T. was supported by the National Science Foundation grants DMS-1818757, DMS-1952757, and DMS-2045646. 

\bibliographystyle{Science}
\bibliography{references}

\setcounter{figure}{0}
\renewcommand{\thefigure}{S\arabic{figure}}
\renewcommand{\thetable}{S\arabic{table}}

\section*{Supplementary Material}
\section{Generating the training data} \label{sec_generation_data}

The training dataset consists of $N$ forcing functions, $f_j:\Omega\to\R$, and associated system responses, $u_j:\Omega\to\R$, which are solutions to the following equation:
\begin{equation} \label{eq_problem_app}
\L u_j = f_j, \qquad \mathcal{D}(u_j,\Omega) = g,
\end{equation}
where $\L$ is a linear differential operator, $\mathcal{D}$ is a linear operator acting on the solutions, $uj$, and the domain, $\Omega$; with $g$ being the constraint. Unless otherwise stated, the training data comprises $N=100$ forcing and solution pairs, where the forcing terms are drawn at random from a Gaussian process, $\mathcal{GP}(0,K_{\text{SE}})$, where $K_{\text{SE}}$ is the squared-exponential covariance kernel~\cite{rasmussen2006gaussian} defined as
\begin{equation} \label{eq_kernel}
K_{\text{SE}}(x,y) = \exp\left(-\frac{|x-y|^2}{2\ell^2}\right),\qquad x,y\in\Omega.
\end{equation}
The parameter $\ell>0$ in \cref{eq_kernel} is called the length-scale parameter, and characterizes the correlation between the values of $f\sim \mathcal{GP}(0,K_{\text{SE}})$ at $x$ and $y$ for $x,y\in \Omega$. A small parameter, $\ell$, yields highly oscillatory random functions, $f$, and determines the ability of the GP to generate a diverse set of training functions. This last property is crucial for capturing different modes within the operator, $\L$, and for learning the associated Green's function accurately~\cite{boulle2021learning}. Other possible choices of covariance kernels include the periodic kernel~\cite{rasmussen2006gaussian}:
\[K_{\text{Per}}(x,y)=\exp\left(-\frac{2\sin^2(\pi|x-y|)}{\ell^2}\right),\qquad x,y\in\Omega,\]
which is used to sample periodic random functions for problems with periodic boundary conditions (\cref{fig_mean}B). Another possibility is a kernel from the Mat\'ern family~\cite{rasmussen2006gaussian}.

\begin{figure}[htbp]
\centering
\vspace{0.5cm}
\begin{overpic}[width=0.8\textwidth]{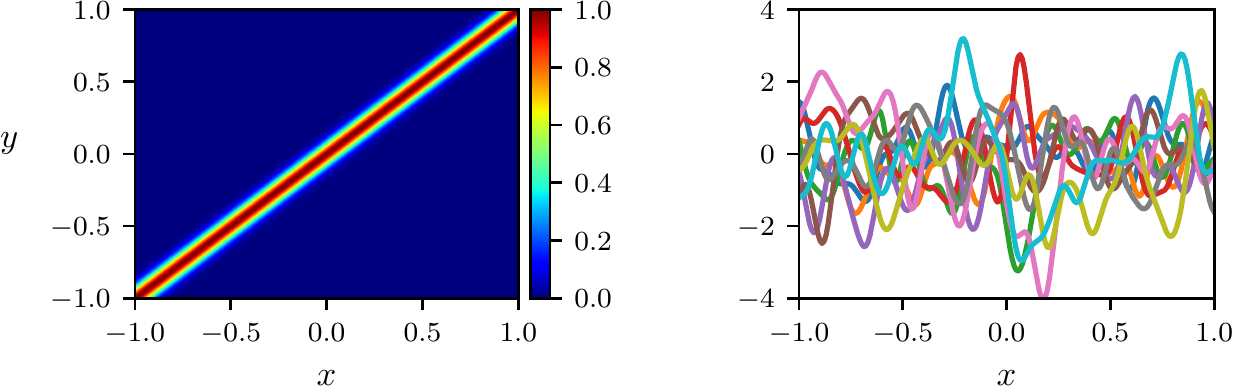}
\put(0,30){\textbf{A}}
\put(54,30){\textbf{B}}
\end{overpic}
\caption{Random forcing terms. Squared exponential covariance kernel $K_{\textup{SE}}$ on $[-1,1]^2$ with normalized length-scale $\lambda=0.03$ (A) together with 10 functions sampled from the Gaussian process $\mathcal{GP}(0,K_{\textup{SE}})$ (B).}
\label{fig_covariance_kernel}
\end{figure}

When $\Omega$ is an interval $[a,b]$, we introduce a normalized length-scale parameter $\lambda = \ell/(b-a)$, so that the method described does not depend on the length of the interval. In addition, we choose $\lambda=0.03$, so that the length-scale, $\ell$, is larger than the forcing spatial discretization size, which allows us to adequately resolve the functions sampled from the GP with the discretization. More precisely, we make sure that $\ell\geq (b-a)/N_f$ so that $1/N_f \leq \lambda$. In \cref{fig_covariance_kernel}, we display the squared-exponential covariance kernel on the domain $\Omega=[-1,1]$, along with ten random functions sampled from $\mathcal{GP}(0,K_{\text{SE}})$.

When~\cref{eq_problem_app} is a boundary-value problem, we generate training pairs by solving the PDE with a spectral method~\cite{trefethen2000spectral} using the Chebfun software system~\cite{driscoll2014chebfun}, written in MATLAB, and using a tolerance of $5\times 10^{-13}$. We also solve the homogeneous problem with zero-forcing, to compare the learned and exact homogeneous solutions. The exact homogeneous solution is not included in the training dataset. When the homogeneous solution is zero, the solutions, $\{u_j\}_{j=1}^{N}$, and forcing terms, $\{f_j\}_{j=1}^N$, are rescaled, so that $\max_{1\leq j\leq N} \|u_j\|_{L^\infty(\Omega)}=1$. By doing this, we facilitate the training of the NNs by avoiding disproportionately small-scale or large-scale data. In the presence of real data, with no known homogeneous solution, one could instead normalize the output of the NNs, $\N_G$ and $\N_{\text{hom}}$, to facilitate the training procedure.

\section{Loss function} \label{sec_loss}

The NNs, $\N_G$ and $\N_{\text{hom}}$, are trained by minimizing a mean square relative error (in the $L^2$-norm) regression loss, defined as:
\begin{equation} \label{eq_loss}
\text{Loss} = \frac{1}{N}\sum_{j=1}^N\frac{1}{\|u_j\|_{L^2(\Omega)}^2}\int_{\Omega} \left(u_j(x) - \mathcal{N}_{\text{hom}}(x) - \int_\Omega \mathcal{N}_G(x,y)f_j(y)\,\textup{d} y\right)^2\,\textup{d} x.
\end{equation}
Unless otherwise stated, the integrals in \cref{eq_loss} are discretized by a trapezoidal rule~\cite{suli2003introduction} using training data values that coincide with the forcing discretization grid, $\{y_i\}_{i=1}^{N_f}$, and measurement points, $\{x_i\}_{i=1}^{N_u}$. As an example, for $1\leq j\leq N$, the squared $L^2$-norm of $u_j$, on a one-dimensional domain $\Omega=[a,b]\subset\R$, is approximated as
\[\|u_j\|_{L^2(\Omega)}^2 = \int_{a}^b u_j(x)^2\d x\approx \sum_{i=2}^{N_u}\frac{u_j(x_{i-1})^2+u_j(x_{i})^2}{2}\Delta_{x_i},\]
where $\Delta_{x_i}=x_i-x_{i-1}$ is the length of the $i$th subinterval $[x_{i-1},x_i]$.

In \cref{sec_location}, we compare the results obtained by using trapezoidal integration, described above, and a Monte-Carlo integration~\cite{binder2012monte}:
\[\|u_j\|_{L^2(\Omega)}^2 \approx \frac{b-a}{N_u}\sum_{i=1}^{N_u}u_j(x_{i})^2,\]
which has a lower convergence rate to the integral with respect to the number of points, $N_u$. This integration technique is, however, particularly suited for approximating integrals in high dimensions, or with complex geometries~\cite{binder2012monte}.

It is also possible to enforce some prior knowledge, regarding the Green's function, through the loss function, by adding a penalty term. If the differential operator is self-adjoint, then depending on the constraint operator $\mathcal{D}$, the associated Green's function is symmetric, \emph{i.e.}, $G(x,y) = G(y,x)$ for all $x,y\in\Omega$. In this case, one can train a symmetric NN $\N_G$ defined as 
\[
\N_G(x,y) = \mathcal{N}(x,y) + \mathcal{N}(y,x), \qquad x,y\in\Omega.
\]
However, our numerical experiments reveal that the NNs can learn both boundary conditions and symmetry properties directly, from the training data, without additional constraints on the loss function or network architectures.

\begin{figure}[htbp]
\centering
\begin{overpic}[width=0.5\textwidth, trim=0 0 0 0,clip]{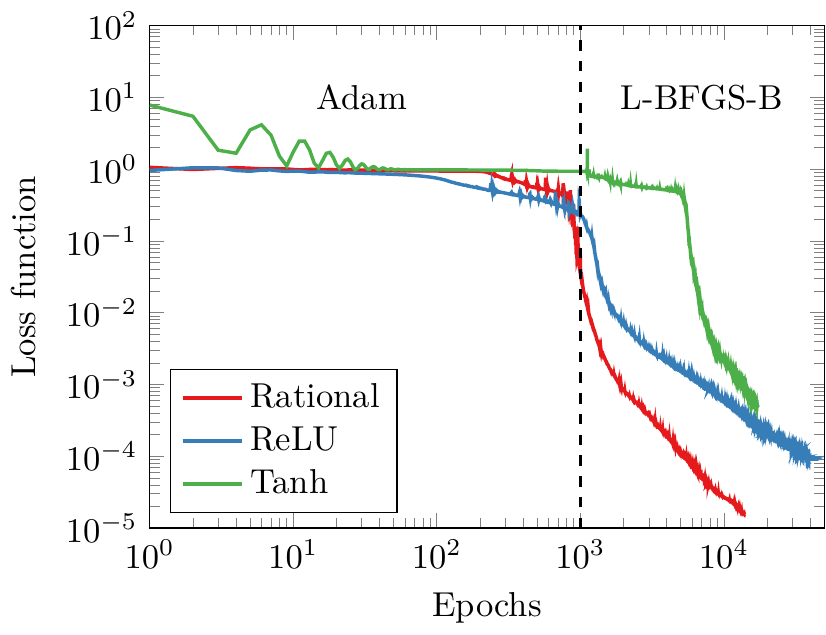}
\end{overpic}
\caption{Loss function during training. Loss function magnitudes for the ReLU, tanh, and rational NNs with respect to the number of epochs. The networks are trained to learn the Green's function of the Helmholtz operator with homogeneous Dirichlet boundary conditions and frequency $K=15$. Adam's optimizer is used until 1000 epochs (before the dashed line) and L-BFGS-B is employed thereafter.}
\label{fig_loss}
\end{figure}

In \cref{fig_loss}, we display the value of the loss function during the training of the NNs with different activation functions: rational, ReLU, and hyperbolic tangent (tanh). In this example, we aim to learn the Green's function of a high-frequency Helmholtz operator with homogeneous Dirichlet boundary conditions on $\Omega = [0,1]$:
\begin{equation} \label{eq_helmholtz}
\L u = \frac{d^2u}{dx^2}+K^2 u,\qquad u(0)=u(1)=0,
\end{equation}
where $K=15$ denotes the Helmholtz frequency. We first remark in \cref{fig_loss} that the rational NN is easier to train than the other NNs, as it minimizes the loss function to $10^{-5}$ with $\approx 15000$ epochs, while a ReLU NN requires three times as many optimization steps to reach $10^{-4}$. We also see that the loss function for the ReLU and rational NN becomes more oscillatory~\cite{bengio2012practical} and harder to minimize before epoch 1000, while it converges much faster after switching to L-BFGS-B. In theory, one could introduce a variable learning rate that improves the behavior of Adam's optimizer~\cite{george2006adaptive,smith2017cyclical}. However, that introduces an additional parameter, which is not desirable in the context of PDE learning. We aim to design an adaptive and easy-to-use method that does not require extensive hyperparameter tuning. We also observe that the tanh NN has a similar convergence rate to the rational NN due to the smoothness of the activation function, but this network exhibits instability during training, as indicated by the high value of the loss function when the optimization terminates. Rational NNs do not suffer from this issue, thanks to the initialization close to a ReLU NN, as can be observed in \cref{fig_loss}, when focusing on the value of the loss function corresponding to the early optimization steps.

\section{Measuring the results} \label{sec_measure_res}

Once the NNs have been trained, we visualize the Green's functions by sampling the networks on a fine $1000\times 1000$ grid of $\Omega\times\Omega$. In the case where the exact Green's function $G_{\text{exact}}$ is known, we measure the accuracy of the trained NN, $\N_G$, using a relative error in the $L^2$-norm:
\begin{equation} \label{eq_rel_error}
\text{Relative Error} = 100\times \|G_{\text{exact}}-\N_G\|_{L^2(\Omega)}/\|G_{\text{exact}}\|_{L^2(\Omega)}.
\end{equation}
Here, we multiplied by 100 to obtain the relative error as a percentage (\%). This illustrates an additional advantage of using a Green's function formulation: we can create test case problems with known Green's functions and evaluate the method using relative error and offer performance guarantees. The standard approaches in the literature often use best-case and worst-case examples as testing procedures and therefore do not guarantee that the solution operator is accurately learned. The ``worst-case'' examples can be misleading if they consist of functions with similar behavior to the forcing terms already included in the training dataset. Furthermore, since the space of possible forcing terms is of infinite dimension, it is not possible to evaluate the trained NNs on all these functions to obtain a true worst-case example.

\section{Robustness of the method} \label{sec_robustness}

We test the robustness of our DL method for learning Green's functions and homogeneous solutions of differential equations, with respect to the number of training pairs, the discretization of the solutions and forcing terms, and the noise perturbation of the training solutions, $\{u_j\}_{j=1}^N$. For consistency, we perform numerical experiments where we learn the Green's function of the Helmholtz operator with parameter $K=15$ and homogeneous Dirichlet boundary conditions (see \cref{eq_helmholtz}). The performance is measured using the relative error in the $L^2$-norm defined in \cref{eq_rel_error} between the trained network, $\N_G$, and the exact Green's function, $G_{\text{exact}}$, whose analytic expression is given by
\[G_{\text{exact}}(x,y) =
\begin{cases}
\frac{\sin(15x)\sin(15(y-1))}{15\sin(15)},\qquad & \text{if } x\leq y,\\
\frac{\sin(15y)\sin(15(x-1))}{15\sin(15)},\qquad & \text{if } x> y,\\
\end{cases}
\]
where $x,y\in[0,1]$.

\subsection{Influence of the activation function on the accuracy} \label{sec_act_accuracy}

We compare the performances of different activation functions for learning the Green's functions of the Helmholtz operator by training the NNs, $\N_G$ and $\N_{\text{hom}}$, with rational, ReLU, and tanh activation functions. The numerical experiments are repeated ten times to study the statistical effect of the random initialization of the network weights and the stochastic nature of Adam's optimizer. The rational NN achieves a mean relative error of $1.2\%$ (with a standard deviation of $0.2\%$), while the ReLU NN reaches an average error of $3.3\%$ (with a standard deviation of $0.2\%$), which is three times larger. Note that the ten times difference in the loss function between ReLU and Rational NNs, displayed in \cref{fig_loss}, is consistent with the factor of three in the relative error since the loss is a mean squared error and $\sqrt{10}\approx 3$. This indicates that the rational neural networks are not overfitting the training dataset. One of the numerical experiments with a tanh NN terminated early due to the training instabilities mentioned in \cref{sec_loss}: achieving a relative error of $99\%$. We excluded this problematic run when comparing the ReLU and rational NN's accuracy, limiting ourselves only to cases where the training was successful. The ReLU and rational NNs did not suffer from such issues and were always successful. The averaged relative error of the tanh NN, over the nine remaining experiments, is equal to $3.9\%$ (with a standard deviation of $1.4\%$), which is slightly worse than the ReLU NN, with higher volatility of the results.

\begin{figure}[htbp]
\centering
\vspace{0.5cm}
\begin{overpic}[width=0.8\textwidth]{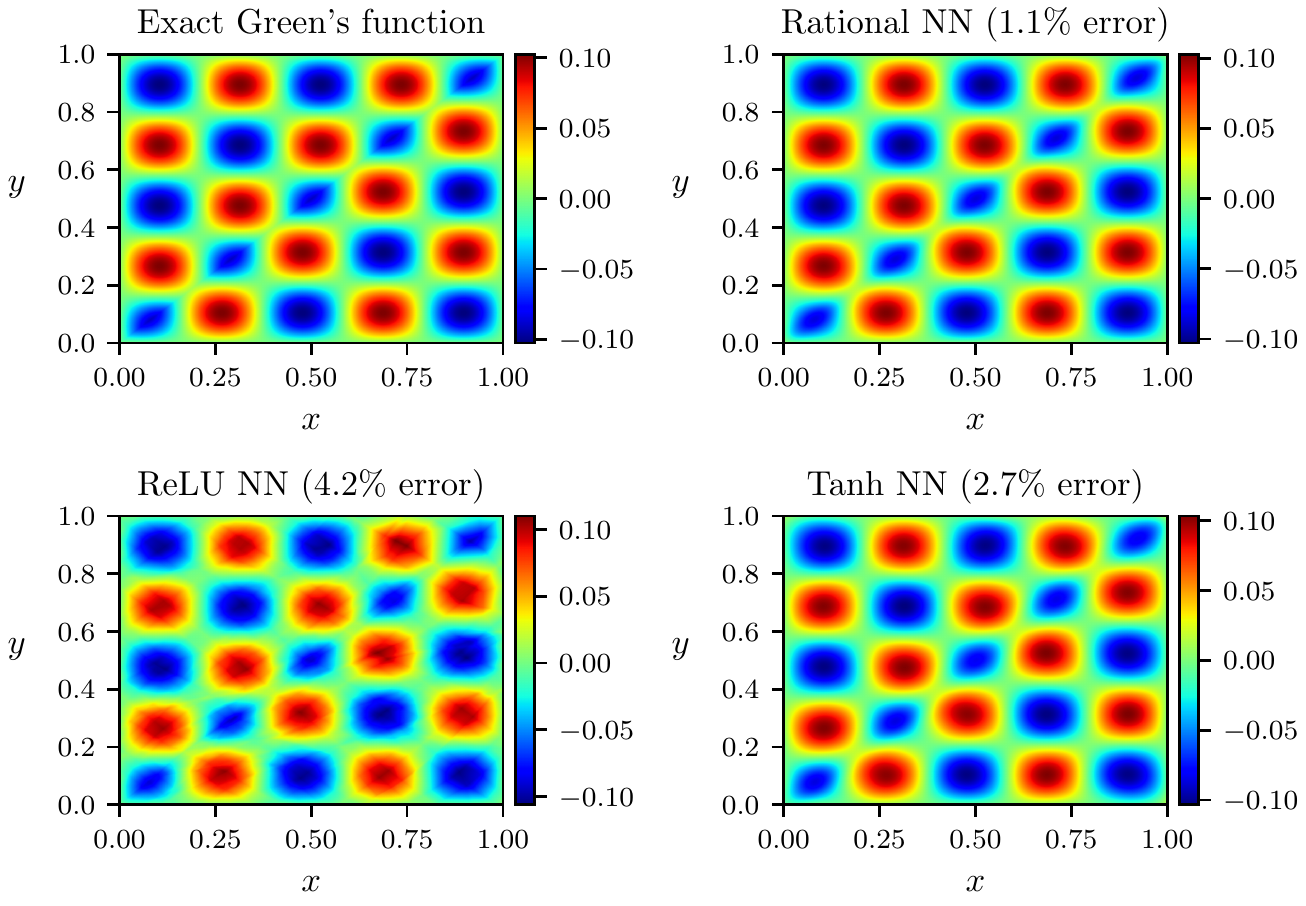}
\end{overpic}
\caption{Comparison of activation functions. Exact and learned Green's functions of the Helmholtz operator by a rational, ReLU, and tanh NN. The relative error in the $L^2$ norm is reported in the titles of the panels.}
\label{fig_activation_functions}
\end{figure}

The exact and learned Green's functions with rational, ReLU, and tanh NNs are displayed in \cref{fig_activation_functions}. We see that the rational and tanh NNs are smooth approximations of the exact Green's function, while visual artifacts are present for the ReLU NN as it is piecewise linear, despite its good accuracy.

\subsection{Number of training pairs and spatial measurements} 

\begin{figure}[htbp]
\centering
\vspace{0.5cm}
\begin{overpic}[width=\textwidth, trim=0 0 0 0,clip]{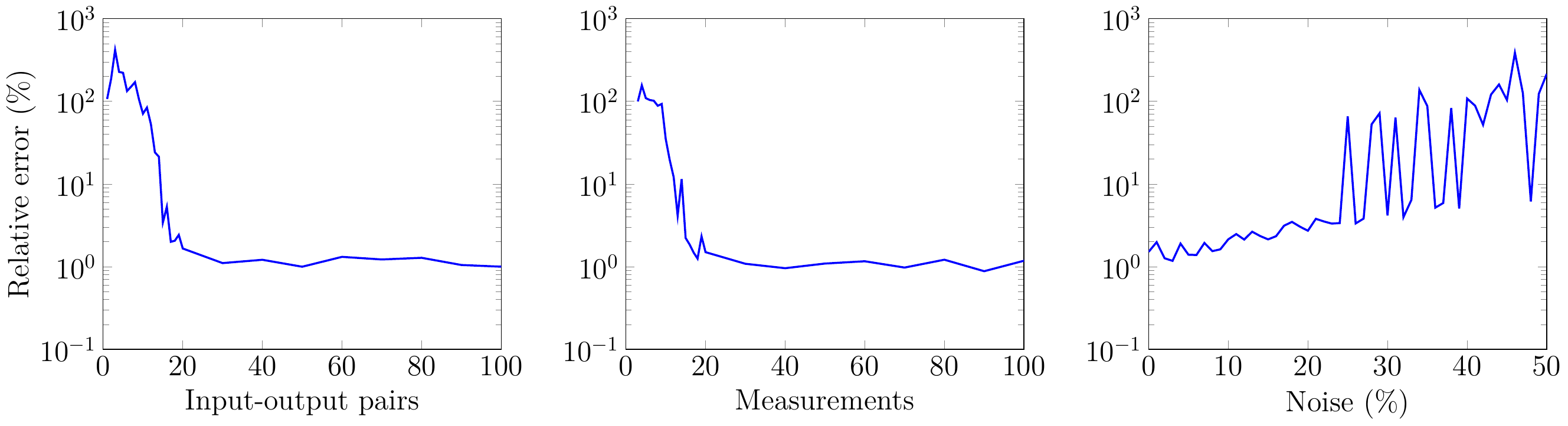}
\put(0,26){\textbf{A}}
\put(34,26){\textbf{B}}
\put(67,26){\textbf{C}}
\end{overpic}
\caption{Robustness of the method. Relative error of the learned Green's function of the Helmholtz operator with respect to the number of input-output pairs (A), spatial measurements (B), and level of Gaussian noise perturbation (C).}
\label{fig_robustness}
\end{figure}

This section describes our method's accuracy as we change the number of training pairs and the size of the spatial discretization. First, we fix the number of spatial measurements to be $N_u=100$, and then vary the number of input-output pairs, $\{(f_j,u_j)\}_{j=1}^N$, of the training dataset for the Helmholtz operator with Dirichlet boundary conditions (see \cref{eq_helmholtz}). As we increase $N$ from $1$ to $100$, we report the relative error of the Green's function learned by a rational NN in \cref{fig_robustness}A. Next, in \cref{fig_robustness}B, we display the relative error on the learned Green's function as we increase $N_u$ from $3$ to $100$, with $N=100$ input-output pairs. Note that we only perform the numerical experiments once since we obtained a low variation of the relative errors in \cref{sec_act_accuracy} when the networks, $\N_G$ and $\N_{\text{hom}}$, have rational activation functions. We observe similar behavior in \cref{fig_robustness}A and B, where the relative error first rapidly (exponentially) decreases as we increase the number of functions in our dataset or spatial measurements of the solutions to the Helmholtz equations with random forcing terms. One important thing to notice is our method's ability to learn the Green's function of a high-frequency Helmholtz operator, with only $1\%$ relative error, using very few training pairs. The performance reaches a plateau at $N\approx 20$ and $N_u\approx 20$, respectively, and ceases to improve. However, the stagnation of the relative error for more numerous training data and spatial measurements is expected and can be explained by our choice of covariance kernel length-scale, which restricts the GP's ability to generate a wide variety of forcing terms. This issue can be resolved by decreasing the length-scale parameter and concomitantly increasing the forcing discretization size or choosing another covariance kernel with a less pronounced eigenvalue decay rate~\cite{boulle2021learning}.

\subsection{Noise perturbation} 

The impact of noise in the training dataset on the accuracy of the learned Green's function is gauged by perturbing the system's response measurements with Gaussian noise as
\begin{equation} \label{eq_noise}
u_j^{\text{noise}}(x_i) = u_j(x_i)(1+\delta c_{i,j}),
\end{equation}
where the coefficients $c_{i,j}$ are independent and identically distributed, Gaussian random variables for $1\leq i\leq N_u$ and $1\leq j\leq N$, and $\delta$ denotes the noise level (in percent). We then vary the level of Gaussian noise perturbation from $0\%$ to $50\%$, train the NNs, $N_G$ and $N_{\text{hom}}$, for each choice of the noise level, and report the relative error in \cref{fig_robustness}C. We first observe a low impact of the noise level on the accuracy of the learned Green's function, as a perturbation of the system's responses measurements with $20\%$ noise only increases the relative error by a factor of $2$ from $1.5\%$ (no noise) to $2.7\%$. When the level of noise exceeds $25\%$, we notice large variations of the relative errors and associated higher volatility in results, characterized by a large standard deviation in error associated with repeated numerical experiments. We consider our DL approach relatively robust to noise in the training dataset.

\subsection{Location of the measurements} \label{sec_location}

As described in the \emph{Methods} and \cref{sec_generation_data}, by default, we use a uniform grid for spatial measurements of the training dataset, and thus we discretize the integrals in the loss function (cf.~\cref{eq_loss}) using a trapezoidal rule. We conducted additional numerical experiments on the Helmholtz example to study the influence of the measurements' location and quadrature rule on the relative error of the learned Green's function. We report the relative errors between the learned and exact Green's functions in \cref{tab_quad}, using a Monte-Carlo or a trapezoidal rule to approximate the integrals and uniform or random spatial measurements. In the latter case, the measurement points $\{x_i\}_{i=1}^{N_u}$ are independently and identically sampled from a uniform distribution, $\mathcal{U}(0,1)$, where $\Omega=[0,1]$ is the domain. We find that the respective relative errors vary between $0.96\%$ and $1.3\%$. Therefore, we do not observe statistically significant differences in the relative error computed by rational NNs. These results support the claim that our method is relatively robust to the type of spatial measurements in the training dataset.

\begin{table}[htbp]
\centering
\caption{Choice of quadrature rules. Relative error of the Green's function of the Helmholtz operator with frequency $K=15$ learned by a rational NN with respect to the type of spatial measurements and quadrature rule (Monte-Carlo or trapezoidal rule) used.}
\label{tab_quad}
\begin{tabular}{c|cc}
\hline
Spatial measurements & Monte-Carlo & Trapezoidal rule \\
\hline
Random & $1.1\%$ & $1.3\%$\\
Uniform & $1.3\%$ & $0.96\%$ \\
\hline
\end{tabular}
\end{table}

\subsection{Missing measurements data} \label{sec_missing_data}

\begin{figure}[htbp]
\centering
\vspace{0.5cm}
\begin{overpic}[width=0.8\textwidth]{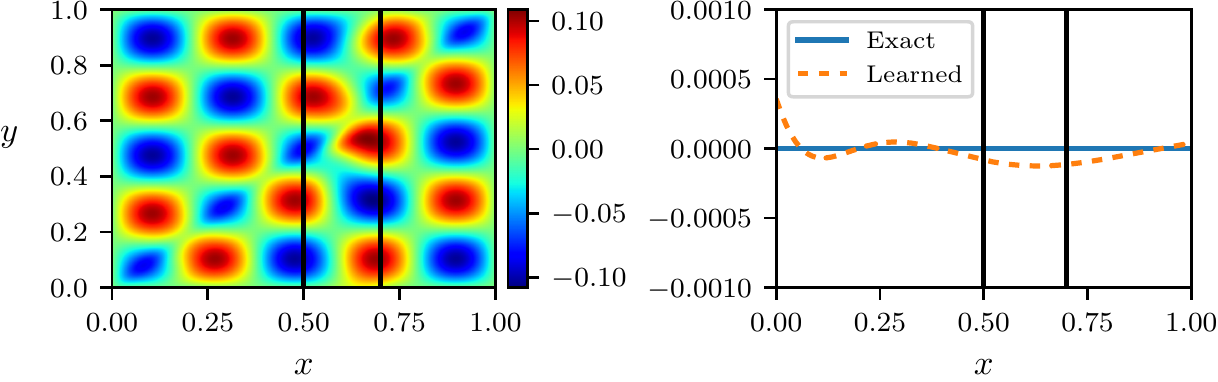}
\put(0,30){\textbf{A}}
\put(51,30){\textbf{B}}
\end{overpic}
\caption{Gap in measurements. (A) Green's function of the Helmholtz operator and its homogeneous solution (B) learned by a rational NN with no measurement points for $x\in[0.5,0.7]$. The space between the vertical black lines indicates where there is a lack of spatial measurements.}
\label{fig_gap}
\end{figure}

Since experimental data may be partially corrupted or unavailable at some spatial locations, we assess our method's accuracy with respect to missing measurement data in the training dataset. We consider the high-frequency Helmholtz operator, defined on the domain $\Omega = [0,1]$ by \cref{eq_helmholtz}, with homogeneous Dirichlet boundary conditions. We introduce a gap in the spatial measurements located at $x\in[0.5,0.7]$ by sampling the system's responses, $\{u_j\}_{j=1}^N$, uniformly on the domain, $[0,0.5]\cup [0.7,1]$. Note that the forcing terms, $\{f_j\}_{j=1}^N$, are still sampled uniformly on the whole domain since they are assumed to be known. The Green's function and homogeneous solution learned by the rational NNs are displayed in \cref{fig_gap}A and B, respectively. Surprisingly, we find that the NN, $\N_G$, can capture the high-frequency pattern of the Green's function and achieves a relative error of $8.2\%$, despite the large gap within the measurement data for $x\in[0.5,0.7]$. Another interesting outcome of this numerical experiment is that the lack of spatial measurements in a specific interval does not influence the accuracy of our method outside this location, \emph{i.e.}, for $x\in [0,0.5]\cup [0.7,1]$ and $y\in [0,1]$ in this example.

\section{Learning features of differential operators from the Green's function} \label{sec_features}

This section highlights that several features of the differential operators can be extracted from the learned Green's function, which supports our aim of uncovering mechanistic understanding from input-output pairs of forcing terms and solutions. 

\subsection{Linear constraints and symmetries} \label{sec_linear_const}

We first remark that boundary constraints, such as the constraint operator, $\D$, of \cref{eq_problem_app}, can be recovered from the Green's function, $G$, of the differential operator, $\L$. Hence, let $f\in C_c^\infty(\Omega)$ be any infinitely differentiable function with a compact support on $\Omega$, and $u$ be the solution to \cref{eq_problem_app} with forcing term, $f$, such that
\[u(x) = \int_{\Omega} G(x,y)f(y)\d y + u_{\text{hom}}(x),\qquad x \in\Omega.\]
Under sufficient regularity conditions, the linearity of the operator, $\D$, implies that $\D(G(\cdot,y),\Omega)=0$ for all $y\in\Omega$. For instance, if $\D$ is the Dirichlet operator: $\D(u,\Omega) = u_{|\partial\Omega}$, then the Green's function satisfies $G(x,y) = 0$ for all $x\in\partial\Omega$.

\begin{figure}[htbp]
\centering
\vspace{0.5cm}
\begin{overpic}[width=0.8\textwidth]{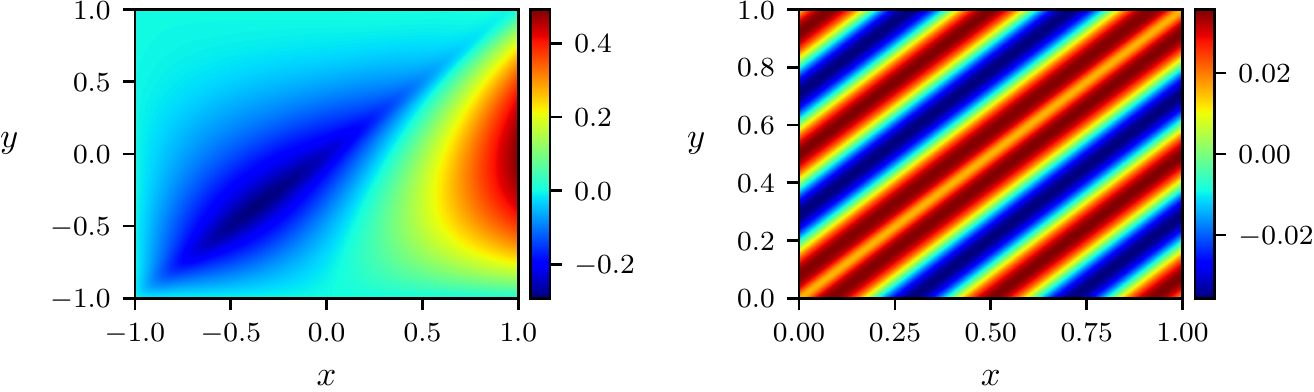}
\put(0,30){\textbf{A}}
\put(50,30){\textbf{B}}
\end{overpic}
\caption{Extraction of linear constraints. (A) Learned Green's functions of a second-order differential operator with an integral constraint defined in \cref{eq_mean_condition}. (B) Green's function of the Helmholtz operator with periodic boundary conditions learned by a rational NN.}
\label{fig_mean}
\end{figure}

As an example, we display in \cref{fig_mean}A the learned Green's function of the following second-order differential operator on $\Omega=[-1,1]$ with an integral constraint on the solution:
\begin{equation} \label{eq_mean_condition}
\L u=\frac{d u^2}{dx^2}+x^2u, \qquad u(-1)=1,\qquad \int_{-1}^1 u(x)\d x=2.
\end{equation}
We observe that $G(-1,y)=0$ for all $y\in[-1,1]$ and one can verify that $\int_{-1}^1 G(x,y)\d x=0$ for any $y\in[-1,1]$. In a second example, we learn the Green's function of the Helmholtz operator on $\Omega=[0,1]$ with frequency $K=15$ and periodic boundary conditions: $u(0)=u(1)$. One can see in \cref{fig_mean}B that the Green's function itself is periodic and that $G(0,y)=G(1,y)$ for all $y\in[0,1]$, as expected. The periodicity of the Green's function in the $y$-direction: $G(x,0)=G(x,1)$ for $x\in[0,1]$, is due to the fact that the Helmholtz operator is self-adjoint, which implies symmetry in the associated Green's function. Furthermore, any linear constraint $\mathcal{C}(u)=0$ such as linear conservation laws or symmetries~\cite{olver2000applications}, satisfied by all the solutions to \cref{eq_problem_app}, under forcing $f\in C_c^\infty(\Omega)$, is also satisfied by the Green's function, $G$, such that $\mathcal{C}(G(\cdot,y))=0$ for all $y\in\Omega$, and is therefore witnessed by the Green's function.

\subsection{Eigenvalue decomposition} \label{sec_eig}

Let $\L$ be a self-adjoint operator and consider the following eigenvalue problem:
\begin{equation} \label{eq_eig_L}
\L v = \lambda v,\qquad \D(v,\Omega) = 0,
\end{equation}
where $v$ is an eigenfunction of the differential operator, $\L$, satisfying the homogeneous constraints with associated eigenvalue, $\lambda> 0$. The eigenfunction, $v$, can be expressed using the Green's function, $G$, of $\L$ as
\[v(x) = \lambda\int_\Omega G(x,y)v(y)\d y,\qquad x\in\Omega,\]
which implies that $v$ is also an eigenfunction of the integral operator with kernel $G$, but with eigenvalue $1/\lambda$. Consider now the eigenvalue problem associated with the Green's function, itself:
\[\int_\Omega G(x,y) w(y)\d y = \mu w(x), \qquad x\in\Omega,\]
where $\mu>0$. Then, we find that $(w,1/\mu)$ are solutions to the eigenvalue problem~\eqref{eq_eig_L}. Consequently, the differential operator, $\L$, and integral operator with kernel, $G$, share the same eigenfunction, but possess reciprocal eigenvalues~\cite{stakgold2011green}. Thus, we can effectively compute the lowest eigenmodes of $\L$ from the learned Green's function.

\begin{figure}[htbp]
\centering
\vspace{0.5cm}
\begin{overpic}[width=0.8\textwidth]{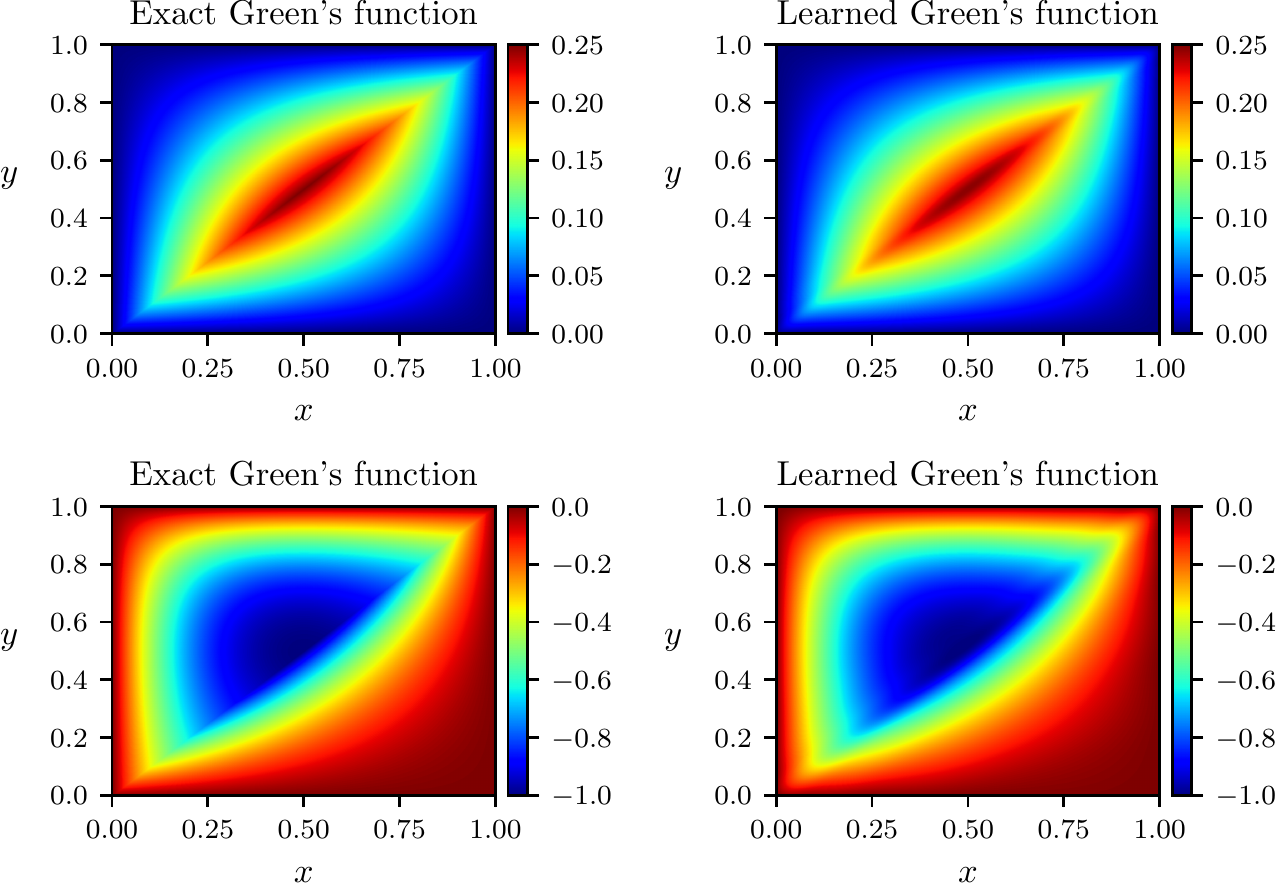}
\put(0,67){\textbf{A}}
\put(0,31){\textbf{B}}
\end{overpic}
\caption{Laplace and advection-diffusion operators. Exact and learned Green's functions of the Laplace (A) and advection-diffusion (B) operators.}
\label{fig_laplace_rational}
\end{figure}

We now evaluate our method's ability to accurately recover the eigenfunctions of the Green's function that are associated with the largest eigenvalues, in magnitude, from input-output pairs. We train a NN to learn the Green's function of the Laplace operator $\L u=-d^2 u/dx^2$ on $[0,1]$, with homogeneous Dirichlet boundary conditions, and numerically compute its eigenvalue decomposition. In \cref{fig_laplace_rational}A, we display the learned and exact Green's function, whose expression is given by 
\[
G_{\text{exact}}(x,y) = 
\begin{cases}
x(1-y), \qquad & \text{if } x\leq y,\\
y(1-x), \qquad & \text{if } y < x,
\end{cases}
\]
for $x,y\in[0,1]$.

\begin{figure}[htbp]
\centering
\vspace{0.5cm}
\begin{overpic}[height=0.56\textwidth, trim=0 0 0 0,clip]{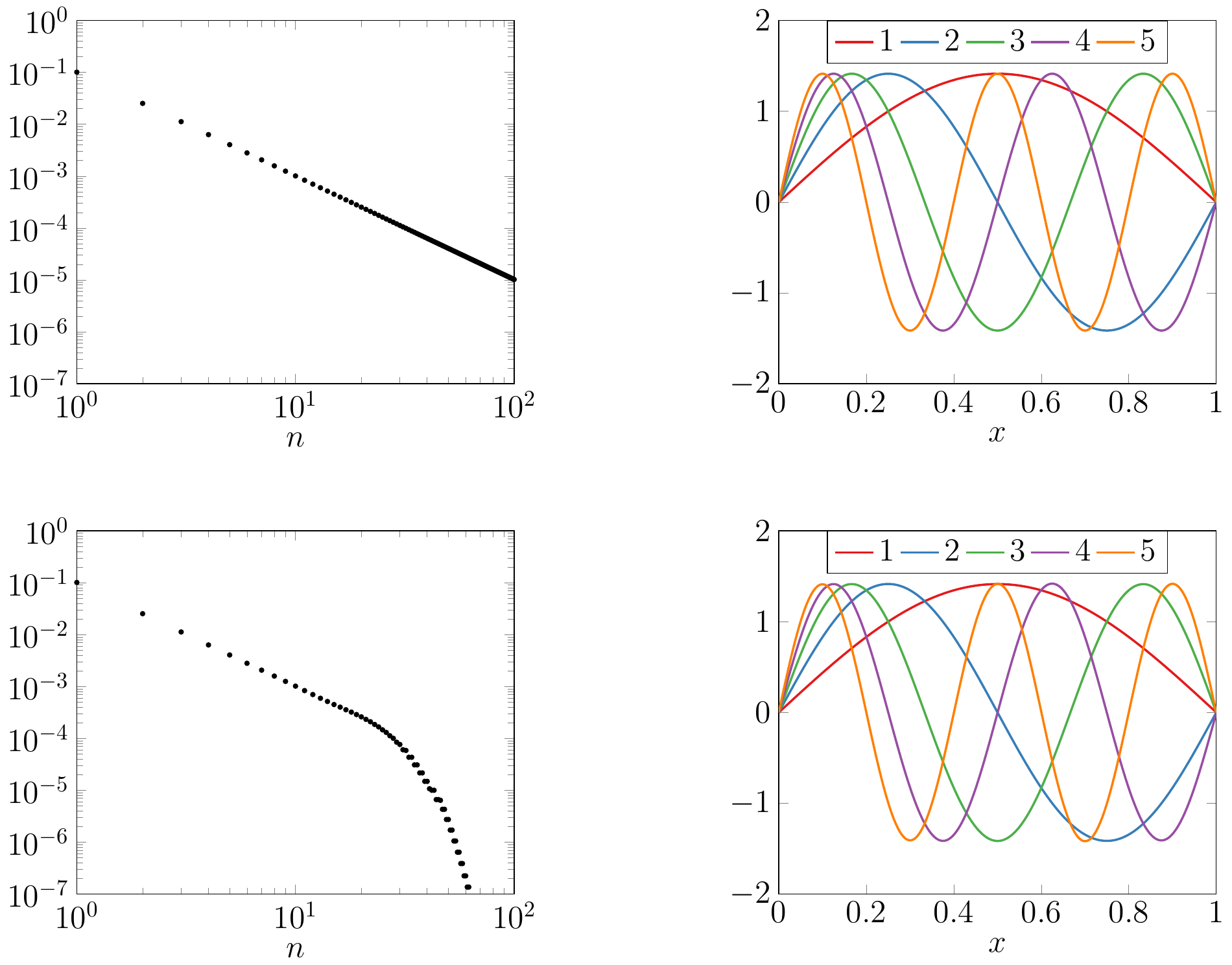}
\put(-5,77){\textbf{A}}
\put(-5,35.5){\textbf{B}}
\put(22.5,80){$D$}
\put(80,80){$V$}
\end{overpic}
\caption{Eigenvalue decomposition. The first 100 largest eigenvalues and first five eigenfunctions of the exact (A) and learned (B) Green's functions of the Laplace operator. The eigenvalues are represented in the left panels, while the right panels illustrate the first five eigenfunctions of the Green's function.}
\label{fig_laplace_eig}
\end{figure}

The one hundred largest eigenvalues in magnitude, along with the corresponding first five eigenfunctions, are visualized for the exact and learned Green's functions in \cref{fig_laplace_eig}. Note that the eigenvectors of the learned Green's functions are normalized and flipped to match the ones of the exact Green's function because eigenfunctions are unique up to a scalar multiple when the eigenvalues are all distinct. We find that we can recover the largest eigenvalues and eigenfunctions of the learned Green's function and that the first 20 largest eigenvalues remain accurate. Therefore, the approximation error between the learned and exact Green's functions mainly affects the smallest eigenvalues. This is an essential feature of our method since the dominant eigenmodes of the differential operator $\L$ are associated with the largest eigenvalues of the Green's functions, which can be learned accurately. The exponential decay of the smallest eigenvalues of the learned Green's function in the left panel of \cref{fig_laplace_eig}B is because the rational NN is a smooth approximation to the exact Green's function.

\subsection{Singular value decomposition}

When the Green's function of the differentiation operator, $\L$, is square-integrable, its associated Hilbert--Schmidt integral operator defined by
\[\mathscr{F}_G f(x) = \int_{\Omega}G(x,y)f(y)\d y,\qquad x\in\Omega,\]
 is compact and admits a singular value decomposition (SVD)~\cite{hsing2015theoretical}. Then, there exist a positive sequence $\sigma_1\geq \sigma_2\geq \cdots> 0$, and two orthonormal bases, $\{\phi_n\}$ and $\{\psi_n\}$, of $L^2(\Omega)$ such that
\begin{equation} \label{eq_svd}
u(x) = \int_{\Omega}G(x,y)f(y)\d y + u_{\text{hom}}(x) = \sum_{n=1}^\infty \sigma_n \langle\phi_n,f\rangle\psi_n(x) + u_{\text{hom}}(x),\qquad x\in\Omega,
\end{equation}
where $u$ is the solution to \cref{eq_problem_app} with forcing term $f$, and $\langle\cdot,\cdot\rangle$ denotes the inner product in $L^2(\Omega)$. Therefore, the action of the solution operator $f\mapsto u$ can be approximated using the SVD of the integral operator. Similarly to \cref{sec_eig} with the eigenvalue decomposition, the dominant terms in the expansion of \cref{eq_svd} are associated with the largest singular values $\sigma_1 \geq \sigma_2\geq \cdots>0$ of the integral operator.

\begin{figure}[htbp]
\centering
\vspace{0.5cm}
\begin{overpic}[height=0.56\textwidth, trim=0 0 0 0,clip]{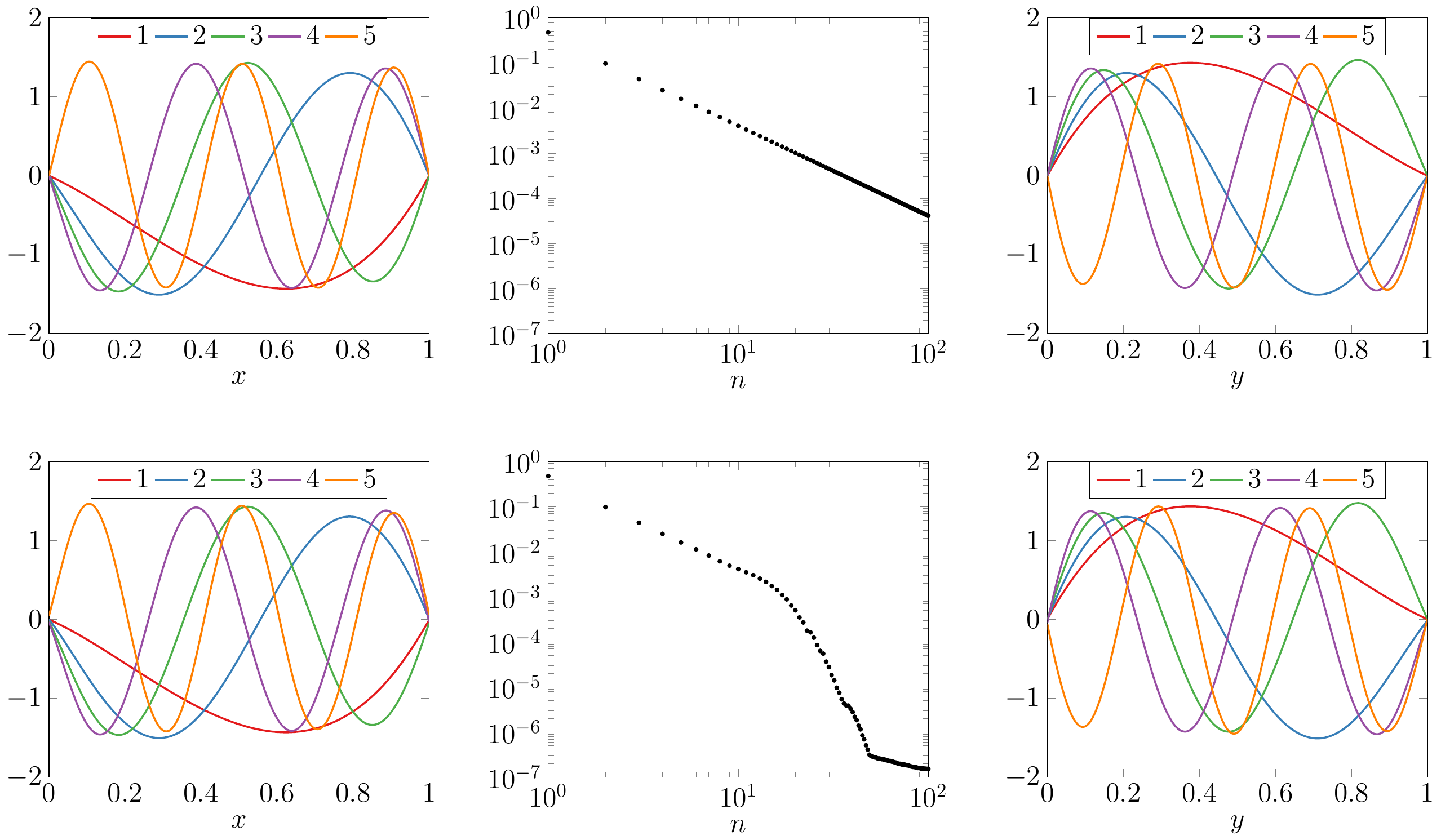}
\put(-3,57){\textbf{A}}
\put(-3,27){\textbf{B}}
\put(15.5,59){$U$}
\put(50,59){$\Sigma$}
\put(85,59){$V$}
\end{overpic}
\caption{Singular value decomposition. Singular value decomposition of the exact (A) and learned (B) Green's functions of the advection-diffusion operator defined by \cref{eq_adv_diff}. The left and right panels, respectively, show the first five left and right singular vectors, $\{\phi\}_{n=1}^5$ and $\{\psi\}_{n=1}^5$, of the exact and learned Green's functions. The singular values of the Green's functions are plotted in the middle panel.}
\label{fig_adv_diff_SVD}
\end{figure}

We now show that one can accurately recover the first singular values and singular vectors from the Green's function learned by a rational NN. We train a rational NN to learn the Green's function of an advection-diffusion operator $\L$ on $\Omega=[0,1]$ with Dirichlet boundary conditions, defined as
\begin{equation} \label{eq_adv_diff}
\L u=\frac{1}{4}\frac{d^2 u}{dx^2}+\frac{du}{dx}+u,\qquad u(0)=1,\, u(1)=-2.
\end{equation}
The learned Green's function is illustrated in \cref{fig_laplace_rational}A, next to the exact Green's function given by:
\[
G_{\text{exact}}(x,y) = 
\begin{cases}
4x(y-1)\exp(-2(x-y)),\qquad & \text{if } x\leq y,\\
(x-1)y, \qquad & \text{if } y < x,
\end{cases}
\]
for $x,y\in[0,1]$. In \cref{fig_adv_diff_SVD}, we display the first five left and right singular vectors and the singular values of the exact and learned Green's functions. We observe that the first fifteen singular values of the learned Green's functions are accurate. This leads us to conclude that our method enables the construction of a low-rank representation of the solution operator associated with the differential operator, $\L$, and allows us to compute and analyze its dominant modes.

\subsection{Schr\"odinger equation with double-well potential}

\begin{figure}[htbp]
\centering
\vspace{0.5cm}
\begin{overpic}[width=0.8\textwidth]{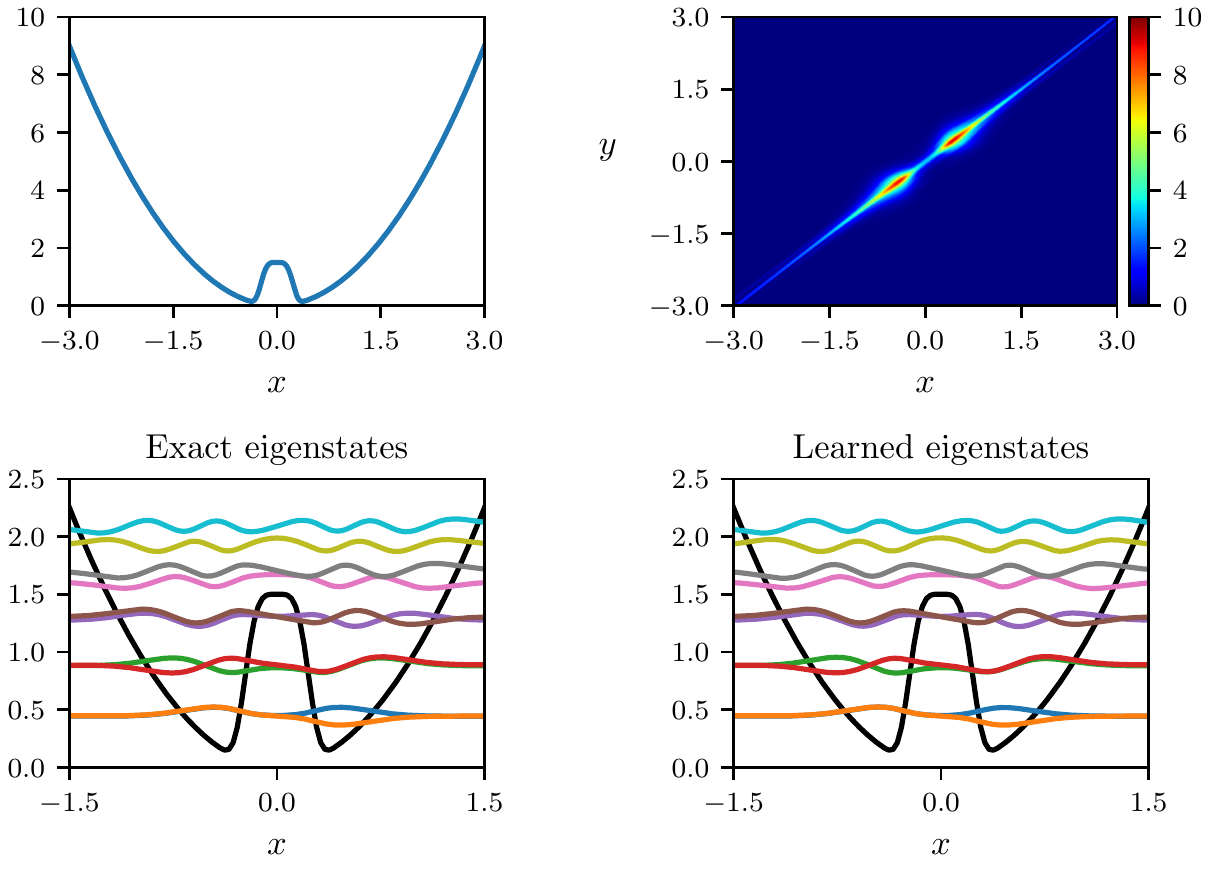}
\put(-3,71){\textbf{A}}
\put(49,71){\textbf{B}}
\put(-3,33){\textbf{C}}
\put(49,33){\textbf{D}}
\end{overpic}
\caption{Schr\"odinger equation. (A) Double well potential $V(x)=x^2+1.5\exp(-(4x)^4)$. (B) Learned Green's function of the Schr\"odinger equation with potential $V(x)$. (C) First ten exact eigenstates computed numerically from the Schr\"odinger operator and (D) eigenstates computed from the learned Green's function displayed in B. The eigenfunctions are shifted by an amount corresponding to the eigenvalue. The double-well potential is shown as a black curve.}
\label{fig_schrodinger_rational}
\end{figure}

We highlight the ability of our DL method to learn physical features of an underlying system by considering the steady-state one-dimensional Schr\"odinger operator on $\Omega=[-3,3]$:
\[\L(u)=-h^2\frac{d^2u}{dx^2}+V(x)u,\qquad u(-3)=u(3)=0,\]
with double-well potential $V(x) = x^2+1.5\exp(-(4x)^4)$ and $h=0.1$~\cite{trefethen2017exploring}. The potential $V(x)$ is illustrated in \cref{fig_schrodinger_rational}, along with the Green's function learned by the rational NN from pairs of forcing terms and the system's responses. First, the shape of the well potential can be visualized along the diagonal of the Green's function in \cref{fig_schrodinger_rational}B. Next, in \cref{fig_schrodinger_rational}, we compute the first ten eigenstates of the Schr\"odinger operator in Chebfun~\cite{driscoll2014chebfun} and plot them using a similar representation as Fig.~6.9 of~\cite{trefethen2017exploring}. Similarly to \cref{sec_eig}, we compute the eigenvalue decomposition of the Green's function learned by a rational NN and plot the eigenstates (shifted by the corresponding eigenvalues) in \cref{fig_schrodinger_rational}. Note that the eigenvalues of the operator and the Green's functions are reversed. We observe a perfect agreement between the first ten exact and learned eigenstates. These energy levels capture information about the states of atomic particles modeled by the Schr\"odinger equation.

\subsection{Singularity location and type} \label{sec_singularity}

The input-output function of a rational NN is a high-degree rational function, which means that it has poles (isolated points for which it is infinite). In rational function approximation theory, it is known that the poles of a near-optimal rational approximant tend to cluster near a function's singularities~\cite{trefethen2021exponential}. The clustering of the poles near the singularity is needed for the rational approximant to have excellent global approximation~\cite{stahl1992best,stahl1993best}. Moreover, the type of clustering (algebraic, exponential, beveled exponential) can reveal the type of singularity (square-root, blow-up, non-differentiable) at that location. This feature of rational approximants is used in other settings~\cite{beyene1999pole}.  

\begin{figure}[htbp]
\centering
\vspace{0.5cm}
\begin{overpic}[width=\textwidth]{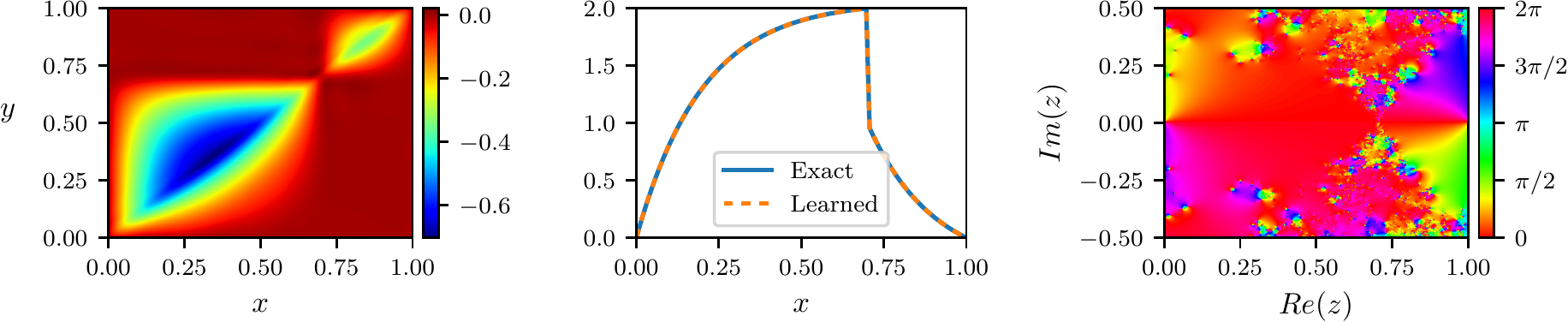}
\put(0,20){\textbf{A}}
\put(34,20){\textbf{B}}
\put(66,20){\textbf{C}}
\end{overpic}
\caption{Singularity location. (A) Learned Green's function of a second-order differential operator with a jump condition at $x=0.7$. Homogeneous solution of the operator with jump condition (B) and argument of the rational NN representing the homogeneous solution in the complex plane (C).}
\label{fig_jump}
\end{figure}

We show that the rational NNs also cluster poles in a way that identifies its location and type. In~\cref{fig_jump}C, we display the complex argument of the trained rational NN for the Green's function of a second-order differential operator with a jump condition, defined on $\Omega=[0,1]$ as
\[\L u = 0.2\frac{d^2 u}{dx^2}+\frac{d u}{dx},\qquad u(0)=u(1)=0,\, u(0.7^{-})=2,\, u(0.7^{+})=1.\]
These diagrams are known as phase portraits and are useful for illustrating complex analysis~\cite{wegert2012visual}. A pole of the rational function can be identified as a point in the complex plane for which the full colormap goes around that point in a clockwise fashion. In  particular, in~\cref{fig_jump}C, we see that the poles of the rational function cluster quite closely to the real-line (where $Im(z) = 0$) at $x = 0.7$. If the clustering is examined more closely, it may be possible to reveal that the singularity in the Green's function at $x = 0.7$ is due to a jump condition. 

\begin{figure}[htbp]
\centering
\vspace{0.5cm}
\begin{overpic}[width=0.8\textwidth]{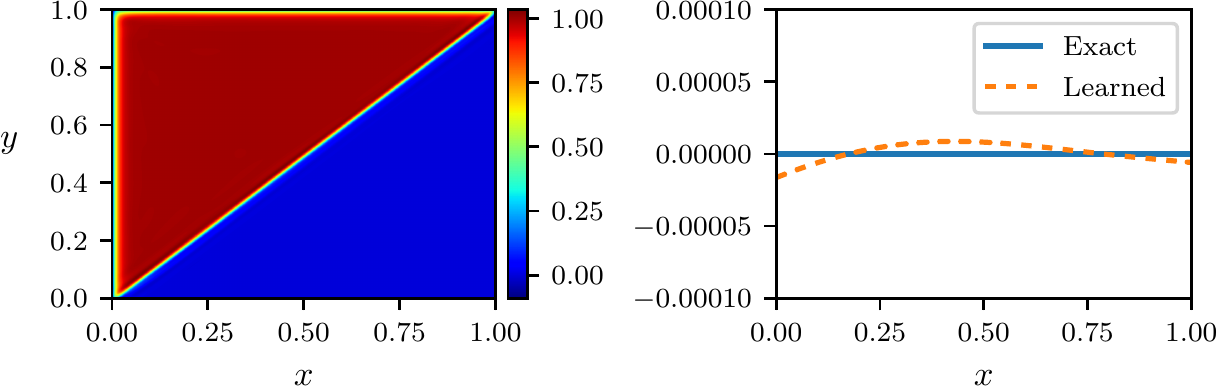}
\put(0,33){\textbf{A}}
\put(51,33){\textbf{B}}
\end{overpic}
\caption{Boundary layer. Learned Green's function (A) and homogeneous solution (B) to a differential equation with a boundary layer around $x=0$.}
\label{fig_boundary_layer}
\end{figure}

Rational NNs are also important for resolving Green's function with boundary layers as the NN can resolve the boundary layer by clustering its poles in the complex plane. In~\cref{fig_boundary_layer}, we see a learned Green's function of a differential equation with a boundary layer at $x = 0$:
\[\L u=-10^{-2}\frac{d^2 u}{dx^2}-\frac{d u}{dx}, \qquad u(0)=u(1)=0, \qquad \Omega=[0,1].\]
While the Green's function is not smooth, our rational NN still resolves it with relatively good accuracy. 

\section{Differential operators in two dimensions} \label{sec_dimension_2}

We demonstrate the ability of our method to learn Green's functions of two-dimensional operators by repeating the numerical experiment of~\cite{li2020neural}, which consists of learning the Green's function of the Poisson operator on the unit disk $\Omega=D(0,1)$, with homogeneous Dirichlet boundary conditions:
\[\L u = \nabla^2 u,\qquad u_{|\partial D(0,1)} = 0.\]
This experiment is a good benchmark for PDE learning techniques as the analytical expression of the Green's function in Cartesian coordinates can be expressed as~\cite{myint2007linear}:
\[G_{\text{exact}}(x,y,\tilde{x},\tilde{y}) = \frac{1}{4\pi}\ln\left(\frac{(x-\tilde{x})^2+(y-\tilde{y})^2}{(x\tilde{y}-\tilde{x}y)^2+(x\tilde{x}+y\tilde{y}-1)^2}\right),\]
where $(x,y),(\tilde{x},\tilde{y})\in D(0,1)$.

\begin{figure}[htbp]
\centering
\vspace{0.5cm}
\begin{overpic}[width=\textwidth]{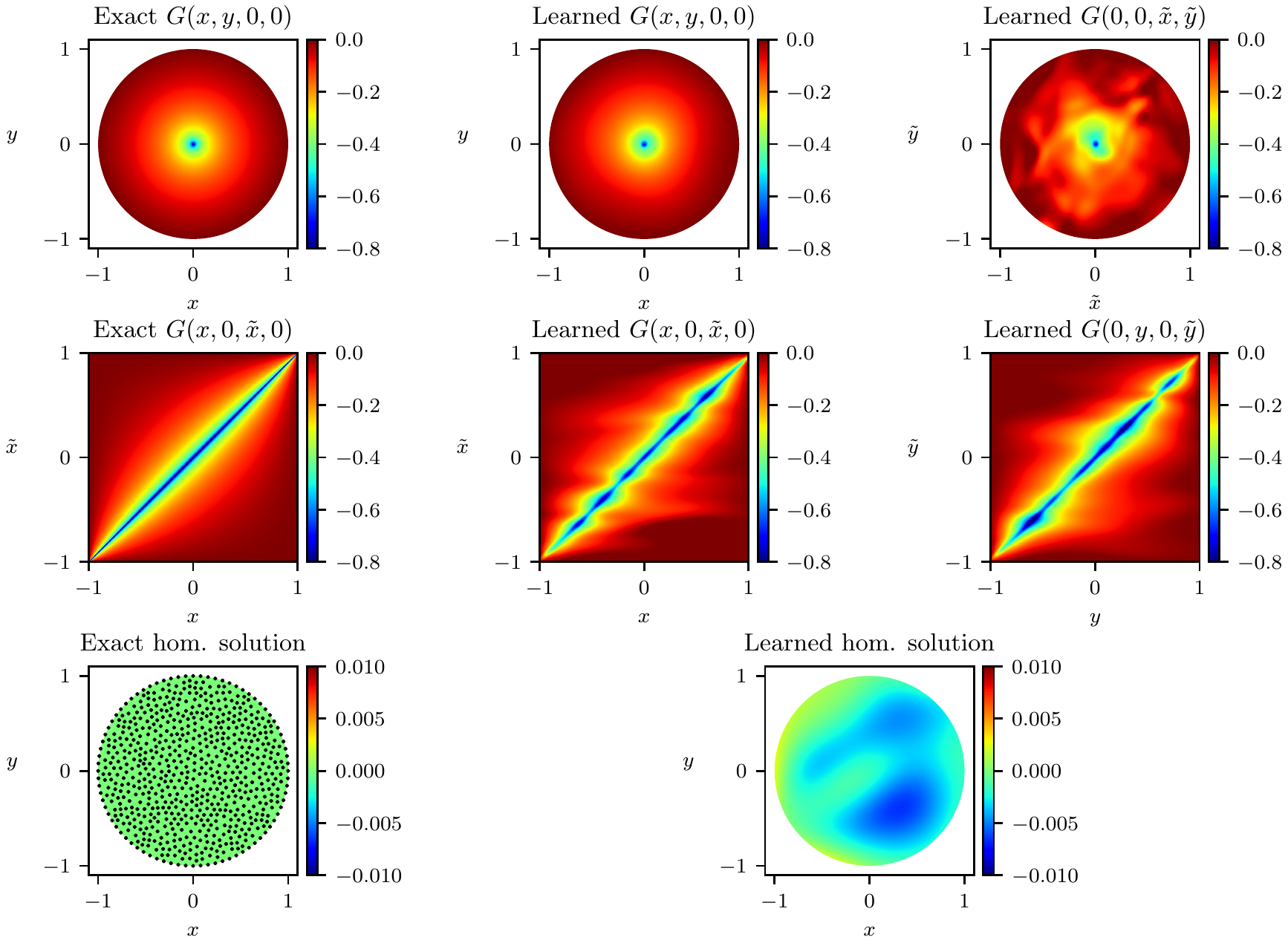}
\put(0,71){\textbf{A}}
\put(34.5,71){\textbf{D}}
\put(69.5,71){\textbf{E}}
\put(0,47){\textbf{B}}
\put(34.5,47){\textbf{F}}
\put(69.5,47){\textbf{G}}
\put(0,23){\textbf{C}}
\put(52.5,23){\textbf{H}}
\end{overpic}
\caption{Poisson equation on the disk. Exact (A and B) and learned (D to F) Green's function of the Poisson operator on the unit disk, evaluated at two-dimensional slices. The colorbar is scaled to remove the singularity of the Green's function at $(x,y)=(\tilde{x},\tilde{y})$. (C) Exact homogeneous solution with sample points for the training functions and (H) homogeneous solution learned by the rational NN.}
\label{poisson_disk}
\end{figure}

The training dataset for this numerical example is created as follows. First, we generate $N=100$ random forcing terms 
using the command \texttt{randnfundisk} of the Chebfun software~\cite{driscoll2014chebfun,filip2019smooth,wilber2017computing} with a frequency parameter of $\lambda = 0.2$, and then solve the Poisson equation, with corresponding right-hand sides, using a spectral method. Then, the forcing terms and system responses (\emph{i.e.}~solutions) are sampled at the $N_u=N_f=673$ nodes of a disk mesh, generated using the Gmsh software~\cite{geuzaine2009gmsh}. The spatial discretization of the mesh is chosen to approximatively match the discretization ($N_u=N_f=625$) of the Stokes example in the main text. Moreover, the mesh structure ensures that the repartition of the sample points is approximately uniform in the disk (\cref{poisson_disk}C) and that the boundary is accurately captured.

The Green's function and homogeneous rational NNs have four hidden layers and width of $50$ neurons, with 4 and 2 input nodes, respectively, as the Green's function is defined on $\Omega\times\Omega$. The two-dimensional integrals of the loss function~\eqref{eq_loss} are discretized using uniform quadrature weights: $w_i = \pi/N_f$ for $1\leq i\leq N_f$. In \cref{poisson_disk}D to G, we visualize four, two-dimensional, slices of the learned Green's function together with two slices of the exact Green's function in panels A and B. Because of the symmetry in the Green's function, due to the self-adjointness of $\L$ and the boundary constraints, the exact Green's function satisfies $G(x,y,0,0) = G(0,0,x,y)$ for $(x,y)\in D(0,1)$. Therefore, we compare \cref{poisson_disk}A to \cref{poisson_disk}D, E, and similarly for \cref{poisson_disk}B and \cref{poisson_disk}F, G. We observe that the Green's function is accurately learned by the rational NN, which preserves low approximation errors near the singularity at $(x,y)=(\tilde{x},\tilde{y})$, contrary to the Neural operator technique~\cite{li2020neural}. The visual artifacts present in \cref{poisson_disk}E to G are likely due to the low spatial discretization of the training data.

\section{System of differential equations} \label{sec_system}

The method for discovering Green's functions of scalar differential operators extends naturally to systems of differential equations. Let $f = \begin{bmatrix}
f^1 & \cdots & f^{n_f}
\end{bmatrix}^\top:\Omega\to\R^{n_f}$ be a vector of $n_f$ forcing terms and 
$u = \begin{bmatrix}
u^1 & \cdots & u^{n_u}
\end{bmatrix}^\top:\Omega\to\R^{n_u}$ be a vector of $n_u$ system responses such that
\begin{equation} \label{eq_system}
\L
\begin{bmatrix}
u^1\\
\vdots\\
u^{n_u}
\end{bmatrix}
=\begin{bmatrix}
f^1 \\ 
\vdots \\
f^{n_f}
\end{bmatrix},\qquad 
D\left(
\begin{bmatrix}
u^1\\
\vdots\\
u^{n_u}
\end{bmatrix},\Omega\right)=
\begin{bmatrix}
g^1\\
\vdots\\
g^{n_u}
\end{bmatrix}.
\end{equation}
The solution to \cref{eq_system} with $f=0$ is called the homogeneous solution and denoted by $u_{\text{hom}}=\begin{bmatrix}
u_{\text{hom}}^1 & \cdots & u_{\text{hom}}^{n_u}
\end{bmatrix}^\top$. Similarly to the scalar case, we can express the relation between the system's response and the forcing term using Green's functions and an integral formulation as
\begin{equation} \label{eq_system_loss}
u^i(x)=
\sum_{j=1}^{n_f}
\int_\Omega G_{i,j}(x,y)f^j(y)\d y+u_{\text{hom}}^{i}(x),\qquad x\in\Omega,
\end{equation}
for $1\leq i\leq n_u$. Here, $G_{i,j}:\Omega\times\Omega\to \R\cup\{\pm\infty\}$ is a component of the \emph{Green's matrix} for $1\leq i\leq n_u$ and $1\leq j\leq n_f$, which consists of a $n_u\times n_f$ matrix of Green's functions:
\[G(x,y) = \begin{bmatrix}
G_{1,1}(x,y) & \cdots & G_{1,n_f}(x,y)\\
\vdots & \ddots & \vdots \\
G_{n_u,1}(x,y) & \cdots & G_{n_u,n_f}(x,y)
\end{bmatrix},\qquad x,y\in\Omega.\]
Following \cref{eq_system_loss}, we remark that the differential equations decouple, and therefore we can learn each row of the Green's function matrix independently. That is, for each row $1\leq i\leq n_u$, we train $n_f$ NNs to approximate the components $G_{i,1},\ldots,G_{i,n_f}$, and one NN to approximate the $i$th component of the homogeneous solution, $u_{\text{hom}}^i$.

\begin{figure}[htbp]
\centering
\vspace{0.5cm}
\begin{overpic}[width=\textwidth]{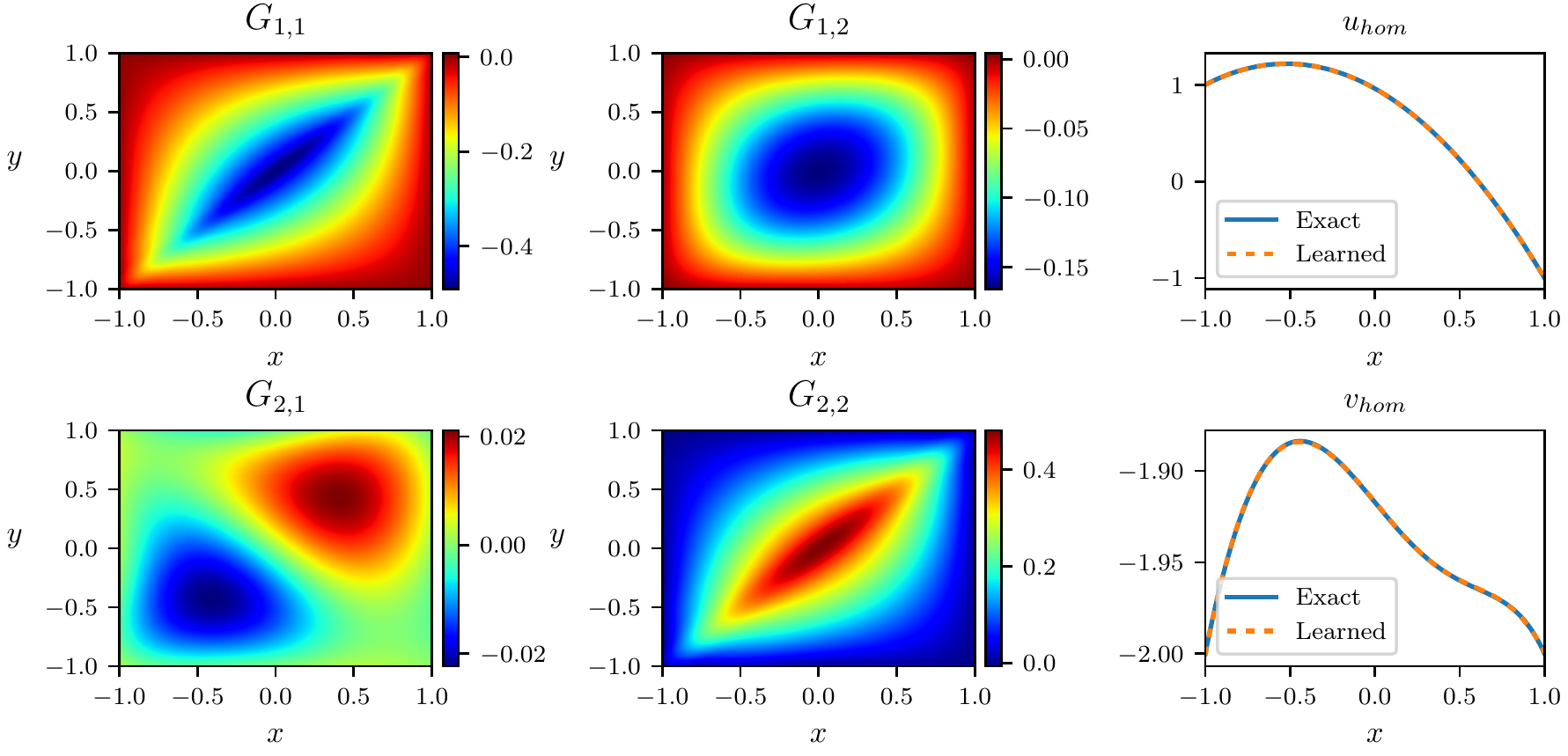}
\put(0,45){\textbf{A}}
\put(72,45){\textbf{B}}
\end{overpic}
\caption{Green's matrix of system of ODEs. (A) Matrix of Green's function learned from the system of ordinary differential equations~\eqref{eq_ODE}. (B) Homogeneous solutions associated with the system of ODEs.}
\label{fig_ODE}
\end{figure}

As an example, we consider the following system of ordinary differential equations (ODEs) on $\Omega = [-1,1]$:
\begin{subequations} \label{eq_ODE}
\begin{align}
\frac{d^2 u}{d x^2} - v &= f^1,\\
\frac{-d^2 v}{d x^2} +xu &= f^2,
\end{align}
\end{subequations}
with boundary conditions: $u(-1) = 1$, $u(1) = -1$, $v(-1)=v(1) =-2$. In \cref{fig_ODE}, we display the different components of the Green's matrix and the exact solution (computed by a spectral method), along with the learned homogeneous solutions. We find that the Green's function matrix provides insight on the coupling between the two system variables, $u$ and $v$, as shown by the diagonal components $G_{1,2}$ and $G_{2,1}$ of the Green's matrix in \cref{fig_ODE}A. Similarly, the components $G_{1,1}$ and $G_{2,2}$ are characteristic of diffusion operators, which appear in \cref{eq_ODE}. In this case, the Green's matrix can be understood as a $2\times 2$ block inverse~\cite{lu2002inverses} of the linear operator, $\L$.

\section{Analysis of main text examples} \label{sec_main_examples}

This section describes the implementation of the main text examples illustrated in Figs.~3 and~4.

\subsection{Viscous shock}

We first consider the following second-order differential operator~\cite{lee1997fast} on $\Omega = [-1,1]$:
\[\L u = 10^{-3}\frac{d^2 u}{dx^2}+2x\frac{du}{dx}, \qquad u(-1) = -1,\, u(1)=1.\]
The robustness of our DL method with missing data in the vicinity of a shock front is analyzed in the panels D to F of Fig.~3, by applying the experiment of \cref{sec_missing_data}. We then remove the measurement points in the interval $[-0.2,0.2]$ and train the Green's function and homogeneous NNs on this new dataset. The trained NNs are sampled on the whole domain to observe the generalization ability of our DL method.

\subsection{Advection-diffusion on the right of the domain}

Fig.~3G to I illustrates the learned Green's function and homogeneous solution to the differential operator:
\[\L u = 0.1\frac{d^2 u}{dx^2}+\mathbb{I}_{(x\geq 0)}\frac{du}{dx},\qquad u(-1)=2,\, u(1)=-1,\]
on $\Omega = [-1,1]$. Here, $\mathbb{I}_{(x\geq 0)}$ denotes the characteristic function on $x\geq 0$. The resulting equation is diffusive on the left half of the domain, while the advection is turned on for $x\geq 0$. We recognize the Green's function of the Laplacian operator, which is responsible for the diffusion behavior, within the visualization of the learned Green's function NN in Fig.~3G, restricted to the domain $[-1,0]\times[-1,0]$. Similarly, the upper right domain of the Green's function is characteristic of an advection-dominated behavior (\cref{fig_boundary_layer}A).
 
\subsection{Linearized models of nonlinear operators} \label{sec_nonlinear}

We emphasize that our DL method can be used to linearize and extract Green's functions from nonlinear boundary value problems of the form
\[\L u+\epsilon\N(u) = f, \qquad \D(u,\Omega) = g,\]
where $\L$ denotes a linear operator, $\N$ is a nonlinear operator, and $\epsilon < 1$ is a small parameter controlling the nonlinearity.
We demonstrate this ability on the three nonlinear boundary value problems, dominated by the linearity, used in \cite{gin2020deepgreen}. 

Fig.~4A of the main text illustrates the learned Green's function of a cubic Helmholtz system on $\Omega=[0,2\pi]$ with homogeneous Dirichlet boundary conditions:
\[\frac{d^2 u}{dx^2}+\alpha u+\epsilon u^3 = f(x),\]
where $\alpha=-1$ and $\epsilon = 0.4$. Next, in Fig.~4B, we consider a nonlinear Sturm--Liouville operator of the form:
\[[-p(x)u']'+q(x)(u+\epsilon u^3) = f(x),\qquad u(0) = u(2\pi) = 0,\]
with $p(x) = 0.4\sin(x)-3$, $q(x)=0.6\sin(x)-2$, and $\epsilon=0.4$. The notation $u'$ denotes the derivative with respect to $x$, $du/dx$. Finally, the example represented in Fig.~4C is the learned Green's function of a nonlinear biharmonic operator:
\[[-p(x)u'']''+q(u+\epsilon u^3) = f(x),\qquad u(0) = u(2\pi) = 0,\]
where $p=-4$, $q=2$, and $\epsilon = 0.4$.

\subsection{Lid-driven cavity problem} \label{sec_lid_driven}

\begin{figure}[htbp]
\centering
\vspace{0.5cm}
\begin{overpic}[width=0.8\textwidth]{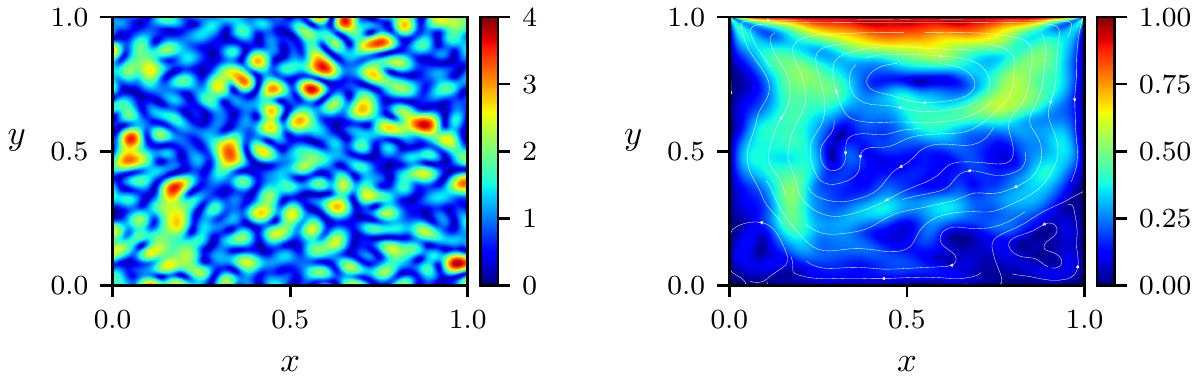}
\put(0,31){\textbf{A}}
\put(51,31){\textbf{B}}
\end{overpic}
\caption{Training functions for Stokes flow. (A) Magnitude of a random applied body force used as a forcing term in the Stokes equations. (B) Velocity magnitude and streamlines of the system's response.}
\label{fig_stokes_training}
\end{figure}

We consider a classical benchmark in fluid dynamics consisting of a two-dimensional lid-driven cavity problem~\cite{elman2014finite}. We aim to discover the matrix of Green's functions of the Stokes flow~\cite{blake1971note}, which is modelled by the following system of equations on the domain $\Omega = [0,1]^2$,
\begin{align*}
\mu\nabla^2\mathbf{u}-\nabla p&=\mathbf{f},\\
\nabla\cdot\mathbf{u}&=0.
\end{align*}
Here, $\mathbf{u}=(u_x,u_y)$ is the fluid velocity, $p$ is the pressure, $\mathbf{f}=(f_x,f_y)$ is an applied body force (\emph{i.e.}~a forcing term), and $\mu=1/100$ is the dynamic viscosity. The fluid velocity satisfies no-slip boundary conditions on the walls, except on the top wall where $\mathbf{u}=(1,0)$. We first generate one hundred forcing terms, $\mathbf{f}$, with two smooth random components, in the Chebfun software~\cite{driscoll2014chebfun,filip2019smooth} using the \texttt{randnfun2} command with wavelength parameter $\lambda = 0.1$. The Stokes equations are then discretized with Taylor--Hood finite elements~\cite{boffi2013mixed,taylor1973numerical} for the velocity and pressure on a mesh with $96\times 96$ square cells and subsequently solved using the Firedrake finite element library~\cite{rathgeber2016firedrake}. We illustrate in \cref{fig_stokes_training} an example of applied body force and velocity solution obtained by solving the system of PDEs. We then create the training dataset for the NNs by sampling the applied body forces and corresponding velocity solutions, $\mathbf{u}$, on a regular $25\times 25$ grid. The four Green's functions and two homogeneous NNs have the same architecture as the one described in the \emph{Methods}, except that they have respectively four and two input nodes (instead of two and one) due to the current spatial dimension.

The four components of the Green's matrix for the Stokes flow are evaluated on the two-dimensional slice $(x,y,0.5,0.5)$, for $x,y\in[0,1]$, and displayed in Fig.~4D of the main text. This figure allows us to visualize the system's response to a point force, $\mathbf{f}=(f_x,f_y)$, located at $(0.5,0.5)$, with the system's response being denoted as $\mathbf{u}=(u_x,u_y)$, where
\begin{align*}
u_x(x,y) &= G_{1,1}(x,y,0.5,0.5)f_x+G_{1,2}(x,y,0.5,0.5)f_y,\\
u_y(x,y) &= G_{2,1}(x,y,0.5,0.5)f_x+G_{2,2}(x,y,0.5,0.5)f_y,
\end{align*}
for $x,y\in[0,1]$. The visualization of the $G_{2,2}$ component in Fig.~4D, corresponding to the system's response to a unitary vertical point force $\mathbf{f}=(0,1)$ is reminiscent of Fig.~1 of~\cite{ekiel2018stokes}. 

\begin{figure}[htbp]
\centering
\vspace{0.5cm}
\begin{overpic}[width=0.7\textwidth]{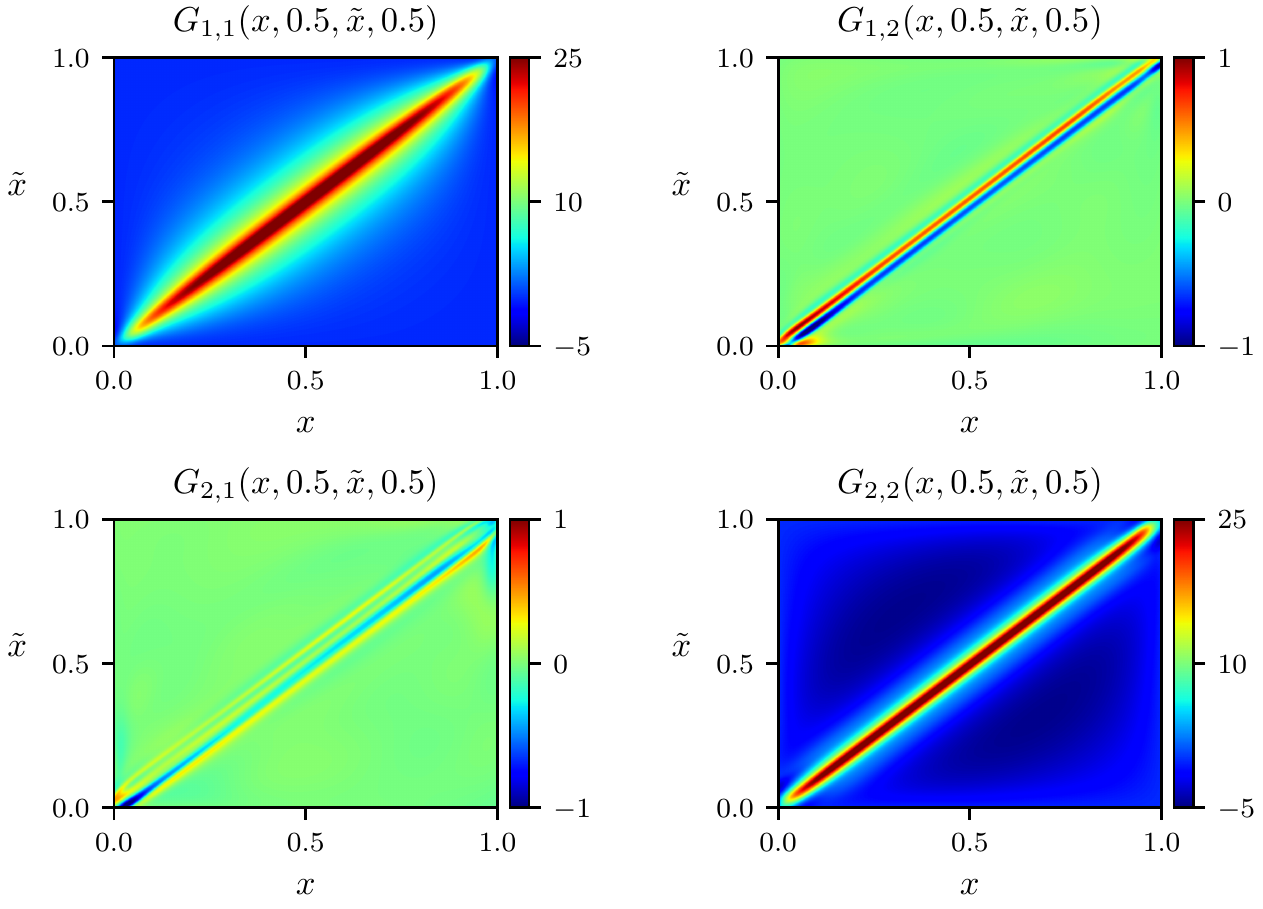}
\end{overpic}
\caption{2nd Green's matrix slice of Stokes flow. The four components of the Green's matrix learned by a rational neural network evaluated at the two-dimensional slice $(x,0.5,\tilde{x},0.5)$.}
\label{fig_slice_stokes_2}
\end{figure}

\begin{figure}[htbp]
\centering
\vspace{0.5cm}
\begin{overpic}[width=0.7\textwidth]{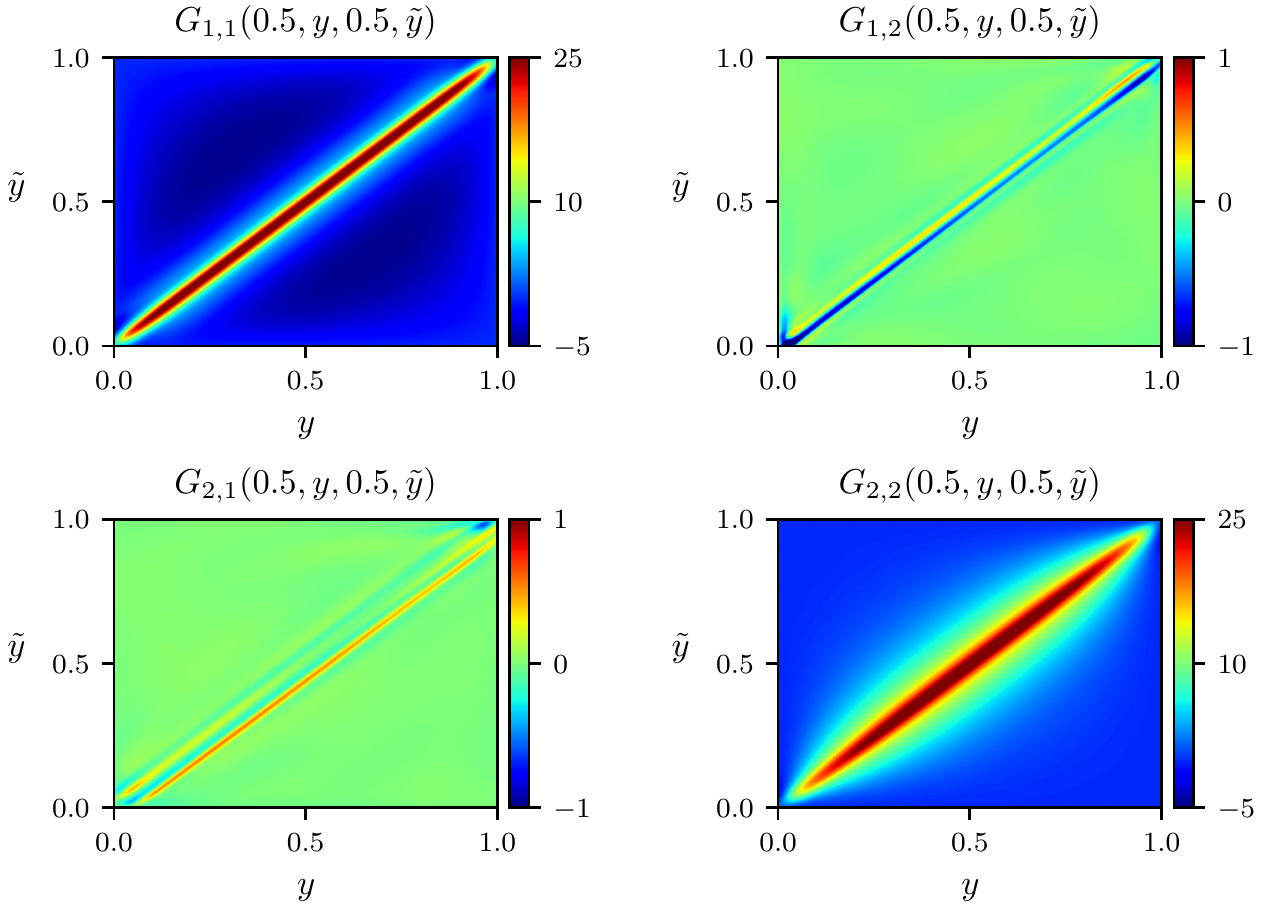}
\end{overpic}
\caption{3rd Green's matrix slice of Stokes flow. The four components of the Green's matrix learned by a rational neural network evaluated at the two-dimensional slice $(0.5,y, 0.5, \tilde{y})$.}
\label{fig_slice_stokes_3}
\end{figure}

\begin{figure}[htbp]
\centering
\vspace{0.5cm}
\begin{overpic}[width=0.7\textwidth]{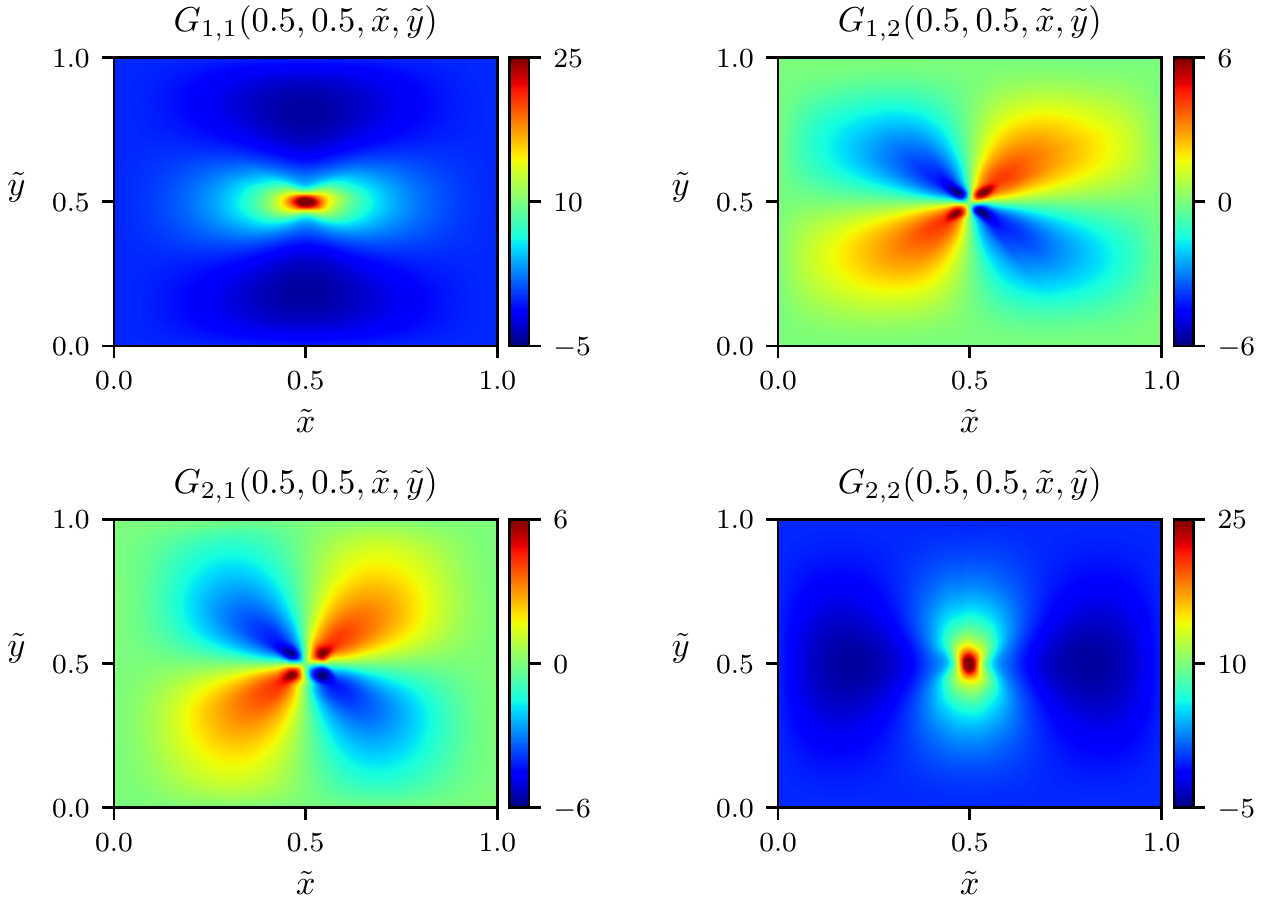}
\end{overpic}
\caption{4th Green's matrix slice of Stokes flow. The four components of the Green's matrix learned by a rational neural network evaluated at the two-dimensional slice $(0.5,0.5,\tilde{x},\tilde{y})$.}
\label{fig_slice_stokes_4}
\end{figure}

Finally, we evaluate the components of the Green's matrix at three other two-dimensional slices: $(x,0.5,\tilde{x},0.5)$, $(0.5,y,0.5,\tilde{y})$, $(0.5,0.5,\tilde{x},\tilde{y})$ and display them respectively in \cref{fig_slice_stokes_2,fig_slice_stokes_3,fig_slice_stokes_4}. These figures illustrate the different symmetries of the Green's matrix, which are captured by the rational NNs. As an example, we see in \cref{fig_slice_stokes_2,fig_slice_stokes_3} that $G_{1,1}(x,0.5,\tilde{x},0.5) = G_{2,2}(0.5,x,0.5,\tilde{x})$ and $G_{2,2}(x,0.5,\tilde{x},0.5) = G_{1,1}(0.5,x,0.5,\tilde{x})$, for $x,\tilde{x}\in [0,1]$. Similarly, we find in \cref{fig_slice_stokes_4} that $G_{1,1}(0.5,0.5,\tilde{x},\tilde{y})=G_{1,1}(0.5,0.5,\tilde{y},\tilde{x})$ and $G_{1,2}(0.5,0.5,\tilde{x},\tilde{y})=G_{2,1}(0.5,0.5,\tilde{x},\tilde{y})$, for $\tilde{x},\tilde{y}\in [0,1]$. The $G_{1,2}$ and $G_{2,1}$ components of the Green's matrix in \cref{fig_slice_stokes_2} highlight a singularity along the diagonal $(x,0.5,x,0.5)$ for $x\in [0,1]$. However, this singularity does not prevent the rational NNs from accurately learning the different components of the Green's matrix displayed in Fig.~4D and \cref{fig_slice_stokes_2,fig_slice_stokes_3,fig_slice_stokes_4}.

\section{Time-dependent equations} \label{sec_time_dep}

In this section, we show that one can use a time-stepping scheme to discretize a time-dependent PDE and learn the Green's function associated with the time-propagator operator $\tau:u_n \to u_{n+1}$, where $u_n$ is the solution of the PDE at time $t = n\Delta t$ for a fixed time step $\Delta t$. As an example, we consider the time-dependent Schr\"odinger equation with a harmonic trap potential $V(x) = x^2$ given by
\begin{equation} \label{eq_schrodinger_time}
i\frac{\partial \psi(x,t)}{\partial t} = -\frac{1}{2}\frac{\partial^2\psi(x,t)}{\partial x^2} + x^2\psi(x,t),\qquad x\in[-3,3],
\end{equation}
with homogeneous Dirichlet boundary conditions. We use a Crank--Nicolson time-stepping scheme with time step $\Delta t = 2\times10^{-2}$ to discretize \cref{eq_schrodinger_time} in time and obtain
\[i\frac{\psi_{n+1}-\psi_{n}}{\Delta t} = \frac{1}{2}\left[-\frac{1}{2}\frac{d^2\psi_{n+1}}{d x^2} + x^2\psi_{n+1}-\frac{1}{2}\frac{d^2\psi_{n}}{d x^2} + x^2\psi_{n}\right].\]
Our training dataset consists of one hundred random initial forcing functions $\psi_n$ at time $t$ and associated response $\psi_{n+1}$ at time $t+\Delta t$. The functions $\psi_n$ have real and imaginary parts sampled from a Gaussian process with periodic kernel and length-scale parameter $\lambda=0.5$ (see \cref{sec_generation_data}), and multiplied by the Gaussian damping function $g(x) = e^{-x^6/20}$ to ensure that the functions decay to zero before reaching the domain boundaries. We then train a rational neural network to learn the Green's function $G$ associated with the time-propagator operator such that
\[\tau(\psi_n)(x) = \int_{-3}^3 G(x,y)\psi_n(y) \d y= \psi_{n+1}(x),\qquad x\in [-3,3].\]
Note that since $\psi$ takes complex values, we in fact split \cref{eq_schrodinger_time} into a system of equations for the real and imaginary parts of $\psi$, and learn the Green's matrix associated with the system (see \cref{sec_system}). 

\begin{figure}[htbp]
\centering
\vspace{0.5cm}
\begin{overpic}[width=\textwidth]{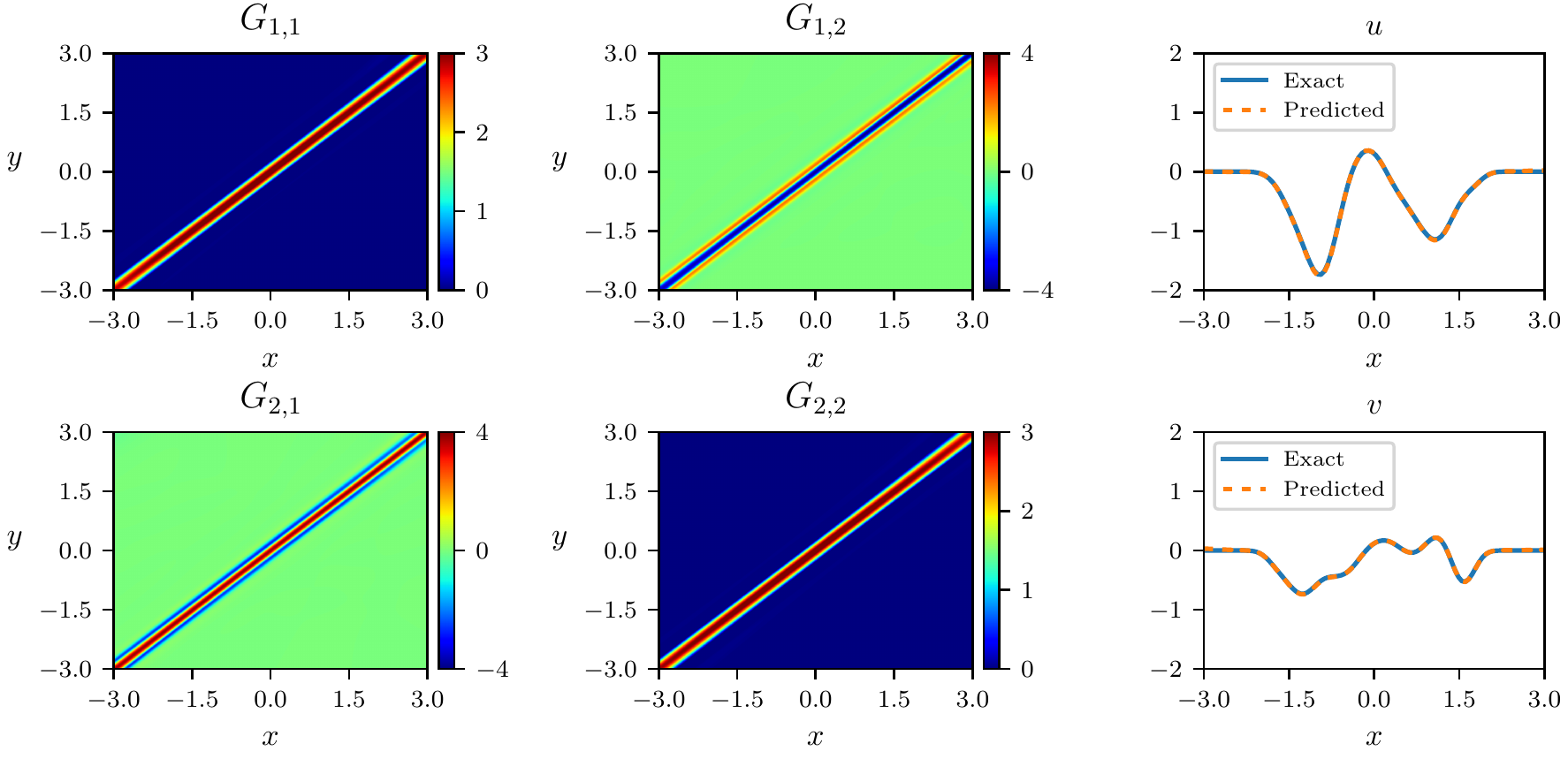}
\put(0,45){\textbf{A}}
\put(70,45){\textbf{B}}
\end{overpic}
\caption{Green's matrix of the time-dependent Schr\"odinger equation. (A) The four components of the Green's matrix for the time propagator operator of the time-dependent Schr\"odinger equation discretized using a time-stepping scheme. (B) Real and imaginary components of the worst case prediction of the solution at the next time step.}
\label{fig_schrodinger_time}
\end{figure}

We report the Green's matrix of the time-propagator operator for the Schr\"odinger equation in \cref{fig_schrodinger_time}A and observe that the four components are dominated by the diagonal, which is expected for a small time-step. Additionally, we evaluate the accuracy of the learned Green's functions by generating a testing dataset with one hundred initial functions $\psi_{n}$, sampled from the same distribution, and associated solution $\psi_{n+1}$ at time $t+\Delta t$. We then compute the average (over the one hundred test cases) relative error in the $L^2$ norm between the exact solution $\psi_{n+1}$ and the one predicted using the learned Green's functions, $\psi_{n+1}^{\text{pred}}$, as
\[\textup{relative error} = \|\psi_{n+1}-\psi_{n+1}^{\text{pred}}\|_{L^2([-3,3])} / \|\psi_{n+1}\|_{L^2([-3,3])},\]
where $\psi_{n+1}^{\text{pred}}$ is defined as
\[\psi_{n+1}^{\text{pred}}(x) = \int_{-3}^3 G(x,y) \psi_n(y)\d y,\qquad x\in[-3,3].\]
Finally, we obtain an average relative error of $1.3\%$ with standard deviation $0.2\%$ across the 100 test cases, confirming the good accuracy of our method. We display the worst-case prediction of the solution $\psi_{n+1}$ in \cref{fig_schrodinger_time}B.

\end{document}